\documentclass[letterpaper]{article} 
\usepackage{aaai25}  
\usepackage{times}  
\usepackage{helvet}  
\usepackage{courier}  
\usepackage[hyphens]{url}  
\usepackage{graphicx} 
\usepackage{natbib}  
\usepackage{caption} 
\usepackage{algorithm}
\usepackage{algorithmic}
\usepackage{enumerate}

\usepackage{subfigure}
\usepackage{amsmath}
\usepackage{breqn}
\usepackage{nicematrix}
\usepackage{multirow}
\usepackage{booktabs}
\usepackage{amssymb}
\usepackage{makecell}
\usepackage{cite}  

\usepackage{appendix} 

\usepackage{newfloat}
\usepackage{listings}
\DeclareCaptionStyle{ruled}{labelfont=normalfont,labelsep=colon,strut=off} 
\floatstyle{ruled}
\newfloat{listing}{tb}{lst}{}
\floatname{listing}{Listing}
\title{To Predict or Not To Predict? \\ Proportionally Masked Autoencoders for Tabular Data Imputation}
\author{
    Jungkyu Kim, 
    Kibok Lee\textsuperscript{\rm*},
    Taeyoung Park\thanks{Co-corresponding authors}
}
\affiliations{
    Department of Statistics and Data Science, Yonsei University, South Korea\\

    \{kim.normal, kibok, tpark\}@yonsei.ac.kr
}

\begin{document}

\maketitle

\begin{abstract}

Masked autoencoders (MAEs) have recently demonstrated effectiveness in tabular data imputation. However, due to the inherent heterogeneity of tabular data, the uniform random masking strategy commonly used in MAEs can disrupt the distribution of missingness, leading to suboptimal performance. To address this, we propose a proportional masking strategy for MAEs. Specifically, we first compute the statistics of missingness based on the observed proportions in the dataset, and then generate masks that align with these statistics, ensuring that the distribution of missingness is preserved after masking. Furthermore, we argue that simple MLP-based token mixing offers competitive or often superior performance compared to attention mechanisms while being more computationally efficient, especially in the tabular domain with the inherent heterogeneity. Experimental results validate the effectiveness of the proposed proportional masking strategy across various missing data patterns in tabular datasets. Code is available at: \url{https://github.com/normal-kim/PMAE}.

\end{abstract}


\section{Introduction} 

Tabular data often contain missing values in real-world scenarios, posing significant challenges for the deployment of machine learning algorithms~\citep{donders2006gentle}.
Inspired by the recent success of masked autoencoders (MAEs) in representation learning across domains such as computer vision (CV)~\citep{he2022masked} and natural language processing (NLP)~\citep{devlin2018bert},
\citet{du2024remasker} proposed adapting MAEs for tabular data imputation.
However, we argue that naively applying the uniform random masking strategy from MAEs to tabular data results in suboptimal performance due to the intrinsic heterogeneity of tabular data.
Unlike images or word tokens, which are relatively homogeneous and semantically invariant to spatial shifts, tabular data are inherently heterogeneous. That is, each column contains distinct information, making spatial shifts meaningless. Such heterogeneity also extends to the distribution of missing values, which can vary across columns. 
Thus, uniform random masking can unintentionally disrupt these distributions by omitting critical variables that are essential for predicting others \citep{wilms2021omitted}, thereby leading to suboptimal performance~\citep{wu2024multimodal}. These challenges underscore the need for a masking strategy specifically designed to account for the heterogeneity of tabular data.

Regarding the architecture of MAEs,
while Transformers~\citep{vaswani2017attention} have shown strong performance,
their self-attention mechanisms typically focus globally on all columns, which can hinder their ability to effectively capture the local group interactions that are often characteristics of tabular data~\citep{yan2023t2g}. This limitation makes Transformers less suited for capturing the complex relationships between columns, particularly in the presence of missing values. In contrast, we argue that MLP-Mixers~\citep{tolstikhin2021mlp} are better suited for the tabular domain, as their fully-connected layers equipped with activation functions are inherently more capable of capturing such multiple group interactions.

We also address the challenge of evaluating imputation performance using a single metric. Tabular data typically consists of discrete categorical variables and continuous numerical variables, each requiring distinct evaluation metrics. 
However, recent studies~\citep{du2024remasker} often rely solely on the root mean square error (RMSE) across all column types, which fails to adequately assess categorical variables.
Specifically, categorical variables are encoded as uniformly distributed values between 0 and 1, despite lacking inherent relative similarities. 
For example, if a categorical variable is encoded as 0, 0.5, and 1, predicting 0 as 0.5 or 1 is equally incorrect, yet 
RMSE penalizes the latter more, leading to skewed evaluations. 
This underscores the need for a more intuitive and appropriate evaluation metric to accurately assess imputation performance in tabular data.
While accuracy is commonly used for categorical variables, it is not directly comparable to RMSE because it is not normalized.
To overcome this, we propose a unified evaluation metric that combines accuracy for categorical variables and the coefficient of determination ($R^2$) for numerical variables.

\begin{figure}[!t]
  \centering
    \includegraphics[width=0.7\linewidth]{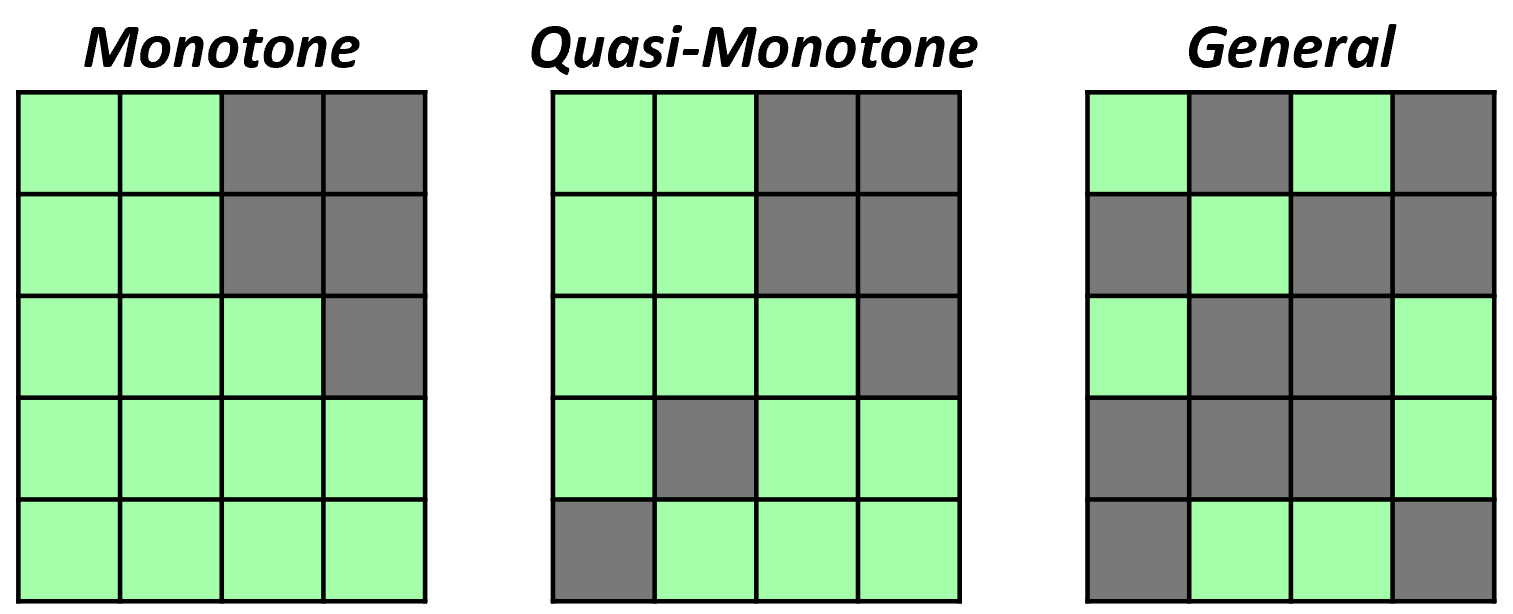}
    \caption{The simplest missing pattern is \textit{Monotone}, where some columns are fully observed, common in longitudinal studies. As the missing pattern becomes \textit{General} (with all columns prone to missing with varying ratios) imputation becomes more challenging. Practitioners need strategies to address these patterns, which are currently under explored. }
    \label{missing_pattern}
  \end{figure}

The main contributions of our work are as follows: 
\begin{itemize}
    \item We propose an MAE with a novel masking strategy based on observed proportions, coined \textbf{P}roportionally \textbf{M}asked \textbf{A}uto\textbf{E}ncoder (\textbf{PMAE}),
    specifically designed to address the challenge of tabular data imputation.
    \item We reveal the effectiveness of MLP-Mixers over Transformers in tabular data imputation tasks through extensive experimental studies.
    \item We demonstrate the efficacy of the proposed \textbf{PMAE} across various missing data distributions and patterns that practitioners commonly encounter.
    Our experimental results indicate that \textbf{PMAE} outperforms the state-of-the-art method~\citep{du2024remasker} by up to 34.1\% in the most challenging \textit{General} pattern (see Figure~\ref{missing_pattern} for an illustration of missing patterns).
\end{itemize}

\section{Related Work}

\subsubsection{Imputation Methods} 
Simple imputation~\cite{hawthorne2005imputing} replaces missing data using summary statistics or KNN-based averages~\cite{troyanskaya2001missing}. Iterative approaches, such as Expectation-Maximization \cite{dempster1977maximum}, MICE~\cite{shah2014comparison}, MissForest~\cite{stekhoven2012missforest}, MIRACLE~\cite{kyono2021miracle}, and Hyperimpute~\cite{jarrett2022hyperimpute}, iteratively refine estimates conditioned on observed data, relying on distributional assumptions. Optimal transport-based approaches, such as TDM~\cite{zhao2023transformed}, impute data in the latent space by matching similar incomplete data batches but may fail to handle categorical data effectively. Generative methods, such as GAIN \cite{yoon2018gain} and MIWAE~\cite{mattei2019miwae}, require fully observed data for initialization, potentially introducing bias. Graph-based approaches, such as IGRM~\cite{zhong2023data}, impute data by modeling sample-wise relationship but do not scale effectively with large sample sizes. ReMasker~\cite{du2024remasker}, which is closely related to our approach, adapts MAE for imputation tasks but applies uniform masking across all columns. This strategy may be less effective for handling complex missing patterns, such as \textit{non-monotone} missing patterns~\cite{sun2018inverse}.

\subsubsection{Propensity Score Weighting Approaches} Our masking function design is inspired by the inverse propensity score weighting method, originally introduced in causal inference \cite{rosenbaum1983central} and later adapted for addressing missing data problems. This approach aims to produce unbiased estimates by using only observed data, assigning higher weights to underrepresented samples \cite{seaman2013review}. However, inaccurate estimation of the propensity score can result in high variance and increased generalization errors \citep{guo2021enhanced,li2023propensity}. While \citet{li2022stabilized} proposed a stabilized approach to mitigate these issues, it requires the joint learning of separate models and extensive parameter tuning, which complicates the optimization process.

\subsubsection{Deep Learning Architectures for Tabular Data} Transformer-based architectures have been widely studied for tabular domain \citep{huang2020tabtransformer, gorishniy2021revisiting, somepalli2021saint, zhang2023mixed}. However, as \citet{yan2023t2g} noted, interactions in tabular data often exist within discrete groups, making soft combinations in self-attention less efficient. Motivated by the effectiveness of the approach proposed by \citet{tolstikhin2021mlp}, we investigate an alternative design to address this issue.


\section{Preliminaries}
\subsubsection{Incomplete Data}
Let $\mathbf{x}_{i}=(x_{i1}, ..., x_{id})^T \in \mathbf{R}^d$ be the $i$-th tabular data with $d$ columns,
sampled from a data distribution $f(\mathbf{x})$.
Without loss of generality on the order of columns, let $\mathbf{x}_i = (\mathbf{x}_{\mathrm{obs}}, \mathbf{x}_{\mathrm{mis}})_i$ be a decomposition, where $\mathbf{x}_{\mathrm{obs}}$ and $\mathbf{x}_{\mathrm{mis}}$ represent the \textit{observed} and \textit{missing} columns of the data, respectively.
For each scalar entry $x_{ij}$, let $\delta_{ij}:= I(x_{ij}$ is observed$)$ be its associated
\textit{missing indicator}.
Then, an incomplete data and its corresponding observed mask are defined as follows:
\begin{enumerate}
    \item The scalar entry of an \textit{incomplete data} is expressed as
        \begin{equation}
            \tilde{x}_{ij} := x_{ij} \cdot I(\delta_{ij} = 1)+\text{nan} \cdot I(\delta_{ij} = 0).
        \end{equation}
    \item The \textit{observed mask} $\mathbf{m}_i \in \mathbf{R}^d$ is the realization of a random missing indicator ${\boldsymbol{\delta}}_i$.
\end{enumerate}

\subsubsection{Missing Mechanism} 
Three types of missing mechanisms commonly occur in the real world \cite{little2019statistical}:
\begin{enumerate}
    \item \textbf{M}issing \textbf{C}ompletely \textbf{A}t \textbf{R}andom (MCAR) occurs when the missingness does not depend on the data, i.e., $\forall{\bf x}$, $P(\boldsymbol{\delta}|{\bf{x}})=P(\boldsymbol{\delta})$.
    \item \textbf{M}issing \textbf{A}t \textbf{R}andom (MAR) occurs when the missingness depends only on the observed data, i.e., $P(\boldsymbol{\delta}|{\mathbf{x}})=P(\boldsymbol{\delta}|\mathbf{x}_{\mathrm{obs}})$.
    \item \textbf{M}issing \textbf{N}ot \textbf{A}t \textbf{R}andom (MNAR) occurs when the missingness does not depend only on the observed data, i.e., $P(\boldsymbol{\delta}|{\bf{x}}) \neq P( \boldsymbol{\delta}|{\mathbf{x}_{\mathrm{obs}}})$.
\end{enumerate}

\subsubsection{Missing Patterns} We propose a categorization of missing data patterns commonly encountered in practice,
as illustrated in Figure~\ref{missing_pattern}. 
When a value $x_{ij}$ is missing for a particular variable $j$, the pattern can be classified as follows:
(i) \textit{Monotone}, 
when there is a rearrangement of columns such that 
all subsequent variables $x_{ik}$ for $k>j$ are also missing for the same observation~$i$ \cite{molenberghs1998monotone},
(ii) \textit{Quasi-Monotone}, which is similar to \textit{Monotone}, but
allowing a few \textit{exceptions} instead of requiring
strict \textit{equality} in the sequence of missing data, and
(iii) \textit{General}, when no specific structure exists.

\subsubsection{Imputation Task} Given an \textit{incomplete dataset} $\mathcal{D}:={\{(\tilde{\mathbf x}_i, \mathbf{m}_{i})\}_{i=1, ..., n}}$, we aim to obtain plausible estimates for inputs $\tilde{\mathbf x}_i$ by learning an imputation function $\hat{f}({\tilde{\mathbf x}_i,\mathbf{m}_i; \hat{\Theta}) }$ that can best approximate the true value of \textit{missing data}: $\hat{x}_{ij} = \hat{f}(\tilde{\mathbf x}_{i}, \mathbf{m}_i; \hat{\Theta}) \cdot I(\delta_{ij} = 0) \approx x_{ij}\cdot I(\delta_{ij}=0)$. 

\section{Motivation}

In this section, we formalize the application of MAEs in the tabular domain.

\subsubsection{Masking Function} 
\begin{figure*}[!ht]
    \centering
        \makebox[\textwidth]{\includegraphics[width=\linewidth]{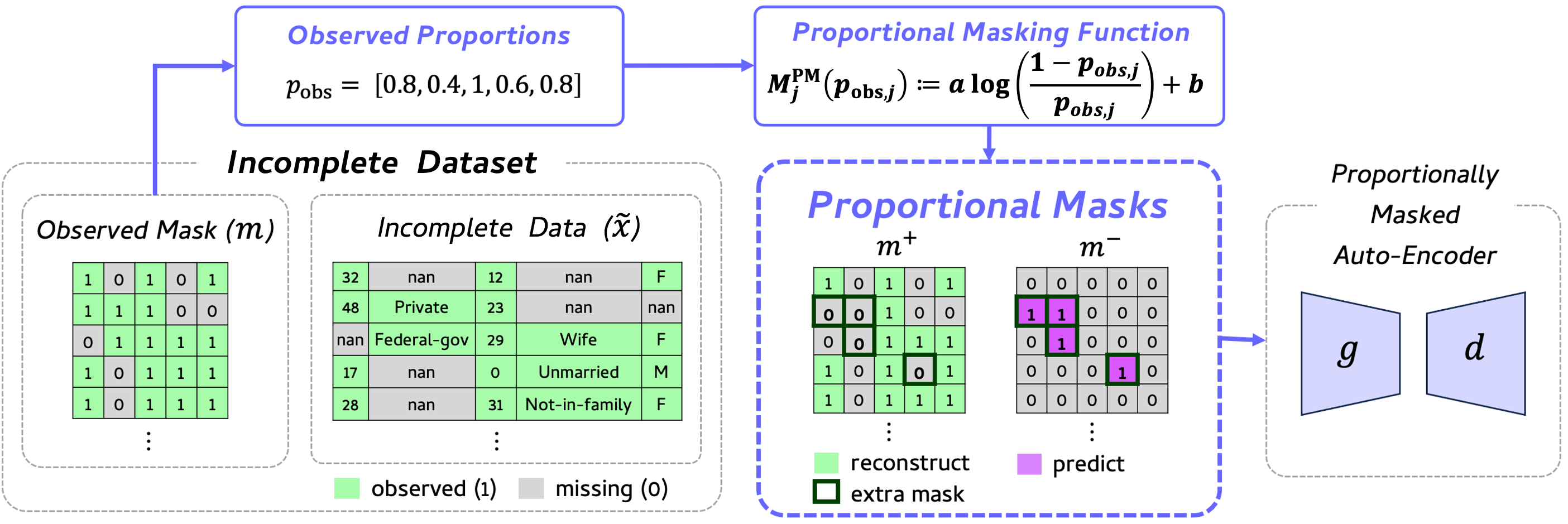}}
    \caption{\textbf{PMAE}. Given the observed mask $\mathbf{m}$, we calculate the observed proportions and apply an additional mask, $\mathbf{m}^-$, where the extra masking probabilities are inversely proportional to the observed proportions. }
    \label{fig:Fig2}
\end{figure*}

Suppose an additional missing mask is applied to the raw data. After this additional masking, the observed mask ${\mathbf{m}}_i$ can be expressed as
\begin{equation}
    \mathbf{m}_i = \mathbf{m}_i^+ + \mathbf{m}_i^-,
\label{eq1}
\end{equation}
where  $\mathbf{m}_i^+$ is an indicator vector representing the parts that \textit{remain observed}
 after applying the additional mask, with entries set to 1 for observed parts, and $\mathbf{m}_i^-$ is an indicator vector with entries set to 1 for \textit{additionally masked} parts (see Fig~\ref{fig:Fig2}).
The process of \textit{additional masking} is as follows: 
\begin{itemize}
    \item Draw a uniform random variable $u_{ij} \sim U(0,1)$, 
    \item Given that $\delta_{ij}=1$, mask according to the following:
    \begin{align}
        m_{ij}^{-} &:= 
        \begin{cases}
            1  & \text{if } u_{ij} < M_j(\cdot),  \\
            0  & \text{otherwise},
        \end{cases}
        \label{masking function}
    \end{align}
\end{itemize}
where $M_j(\cdot) \in \mathbb{R}$ denotes the masking function applied to the column $j$, indicating the extent to which the additional parts of data is masked.

\subsubsection{MAE} A masked autoencoder, $h = d \circ g$, is an encoder-decoder architecture designed to predict the entries with the \textit{additional missing mask}, $\mathbf{m}_i^-$, applied to the raw data. It operates on partial input information, ${\mathbf{x}}'_i = {\mathbf{x}}_i \odot {\mathbf{m}}_i^+$, which is provided due to masking.


\subsubsection{Tabular MAE Loss} While typical MAE in CV/NLP focus on prediction tasks, reconstruction is also important in the tabular domain as they are semantically complex and non-redundant \cite{du2024remasker}. We formally define the \textit{general loss} for the standard MAE of \textit{tabular data} for a sample $i$ in an arbitrary batch $B$ of the $j$-th column as:
\begin{align}
    l_{ij} := \frac{((h(\tilde{\mathbf{x}}_{i} \odot \mathbf{m}_{i}^+))_{j}-x_{ij})\cdot m_{ij})^2 }{\sum_{i\in B} m_{ij}},
    \label{main_loss}
\end{align}
where the $j$-th column of $\tilde{\mathbf{x}}_{i}$
is masked or unmasked based on the value of $m_{ij}^{-}$ in~\eqref{masking function}. 

Since randomness exists in the encoder input through $\mathbf{m}_i^+$ over all $j$, we begin our analysis by fixing all but column $j$. Then, along with $\mathbf{m}_i^+$, we focus on the randomness in the $j$-th entry of the following expression:
\begin{equation}
    \mathbf{m}_i^{0} :=  \mathbf{m}_{i}^{+} - m_{ij}^{+} \mathbf{e}_{j},
\end{equation}
where $\mathbf{e}_{j}$ is the one-hot vector. Let $l_{ij}^0 := \frac{((h(\tilde{\mathbf{x}}_{i} \odot \mathbf{m}_i^{0}))_{j}-x_{ij})^2}{\sum_{i\in B} m_{ij}}$ be a \textit{prediction} loss and $l_{ij}^+ := \frac{((h(\tilde{\mathbf{x}}_{i} \odot \mathbf{m}_i^{+}))_{j}-x_{ij})^2}{\sum_{i\in B} m_{ij}}$ be the \textit{reconstruction} counterpart. Then,~(\ref{main_loss}) can be re-written as: 
\begin{equation}
l_{ij} := m_{ij}^-l_{ij}^{0} + m_{ij}^+l_{ij}^+.
\label{eq.6}
\end{equation}
Note that the MAE will solve for \textit{prediction} task if $m_{ij}^- = 1$ and \textit{reconstruction} task if $m_{ij}^- = 0$. 
Now, since the randomness only exists for $m_{ij}^-$ (or equivalently for $m_{ij}^+ = 1 - m_{ij}^-$, where observed mask $m_{ij} = 1$, if $\delta_{ij} = 1$), we can compute the expectation of the loss for column $j$ defined in (\ref{eq.6}): 
\begin{align}
    \mathbb{E}_{m_{ij}^-}[l_{ij}] 
    &= \mathbb{E}_{m_{ij}^-}[m_{ij}^{-}]l_{ij}^0 + \mathbb{E}_{m_{ij}^-}[m_{ij}^+]l_{ij}^+ \nonumber \\ &= M_j(\cdot)l_{ij}^0 + (1-M_j(\cdot))l_{ij}^+. \label {eq:7}
\end{align}

\subsubsection{The Impact of Uniform Random Masking}  Applying uniform random masking at a constant ratio (i.e., $ M_j(\cdot) \overset{\forall j}{=} 0.5$) in (\ref{eq:7}) essentially assigns equal prediction importance to all columns. However, since tabular data are heterogeneous and exhibit complex relationships between columns, such equal weighting may not be ideal. Moreover, this approach may mask fully observed columns just as likely as partially observed ones. If the inadvertently masked entries are critical for predicting values in other columns, this could lead to \textit{omitted variable bias} \citep{wilms2021omitted}, leading to suboptimal model performance~\citep{wu2024multimodal}. For columns without any missing data, it may be more effective to avoid masking entirely and, consequently, refrain from predicting values that are already fully observed.

\subsubsection{Balancing with Inverse Propensities}  
Moreover, naively computing the loss over only the complete cases may lead to biased estimation. Let us focus on the randomness in the missingness of $\mathbf{x}$; recall that the random variable $\delta_{ij}$, indicates whether $x_{ij}$ is observed. Consider the naive empirical loss for the $j$-th column, $\hat{l}^{\text{naive}}_j := \frac{1}{|B|}\sum_{i\in B} \delta_{ij}\cdot l_{ij}$. Then, 
\begin{align}
\mathbb{E}_{\delta}\left[\hat{l}^{\text{naive}}_j\right] 
&= \mathbb{E}_{\delta}\biggl[\frac{1}{|B|}\sum_{i \in B} l_{ij} - (1-\delta_{ij})\cdot l_{ij} \biggr] \nonumber \\ 
&= \hat{l}^*_j - \mathbb{E}_{\delta}\biggl[\frac{1}{|B|}\sum_{i \in B} (1-\delta_{ij})\cdot l_{ij}\biggr].
 \label{eqn:8}
\end{align}
Since $\delta$ is a random variable that may depend on $\mathbf{x}$, the second term in the last equality generally depends on $\mathbf{x}$ unless $\delta \not\!\perp\!\!\!\perp \mathbf{x}$. Consequently, $\hat{l}^{\text{naive}}_j$ may be a biased estimator of $l^*_j$ due to the second term in~(\ref{eqn:8}).

However, if we have access to the probability of an entry being observed given the realized data, or the \textit{propensity score function}, $\pi_{ij}(\mathbf{x}_i; \boldsymbol{\phi}) := P(\delta_{ij} =1|\mathbf{x}_i; \boldsymbol{\phi})$, we can balance the loss, using only the complete cases. 
Let $\hat{l}^{\text{IPS}}_j := \frac{1}{|B|}\sum_{i\in B}\frac{\delta_{ij}\cdot l_{ij}}{\pi_{ij}(\mathbf{x}_i;\boldsymbol{\phi}^*)}$ be the empirical loss weighted by the inverse propensity score. Then, 
\begin{align}
    \mathbb{E}_{\delta} \left[\hat{l}^{\text{IPS}}_j\right] 
    &= \frac{1}{|B|} \sum_{i \in B} \mathbb{E}_{\delta} \left[ \frac{\delta_{ij}}{\pi_{ij}(\mathbf{x}_i;\boldsymbol{\phi}^*)} \right] l_{ij} \nonumber \\ 
    &= \frac{1}{|B|} \sum_{i \in B} l_{ij} = \hat{l}^*_j .
    \label{ips}
\end{align}
Intuitively, this approach assigns higher importance to less frequently observed samples to balance the overall loss \citep{seaman2013review, kim2021statistical}. 
This ensures that the model adequately accounts for underrepresented samples, which might otherwise have a minimal impact on the overall loss.

However, the practical application of this approach relies on accurately estimating the propensity score function, $\pi_{ij}(\mathbf{x}_i; \boldsymbol{\phi})$. Incorrect estimation could result in unbounded loss values, compromising model performance \cite{li2023propensity}.
To address this challenge, we propose a more implicit masking strategy guided by the following core principles: (i) the \textit{design of the masking function} determines which samples are assigned higher prediction loss, (ii) masks should be applied in proportions that align with the statistical characteristics of the dataset to avoid omitted variable bias, and (iii) as shown in (\ref{ips}), the loss can be balanced by assigning weights inversely to the observed proportions.

\begin{figure}[!t]
    \centering
        \subfigure[Proportional masking]{\includegraphics[width=0.42\linewidth]{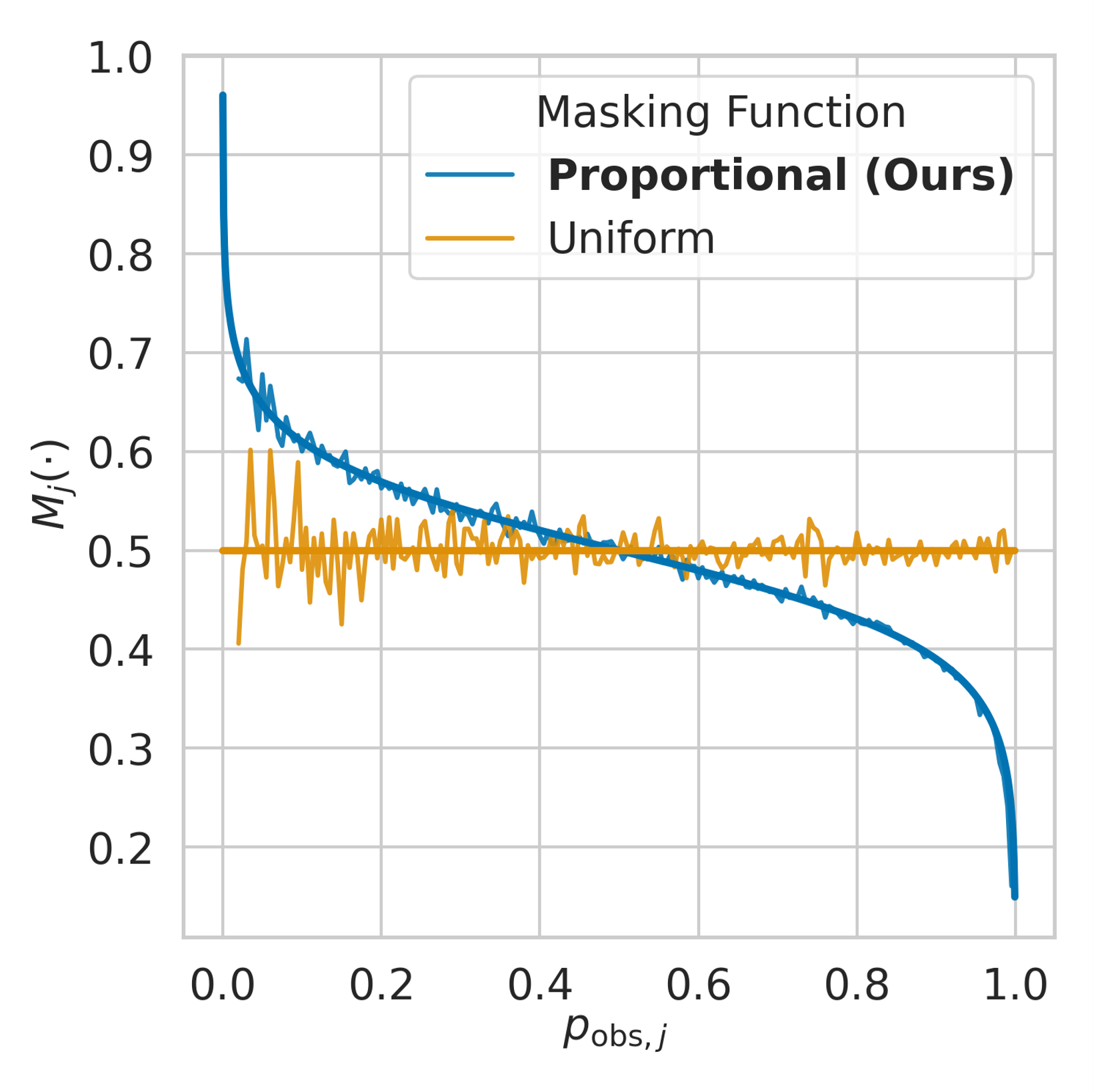}} \hfill%
    \subfigure[Comparison of $\mathbf{m}^+$ after masking]{\includegraphics[width=0.56\linewidth]{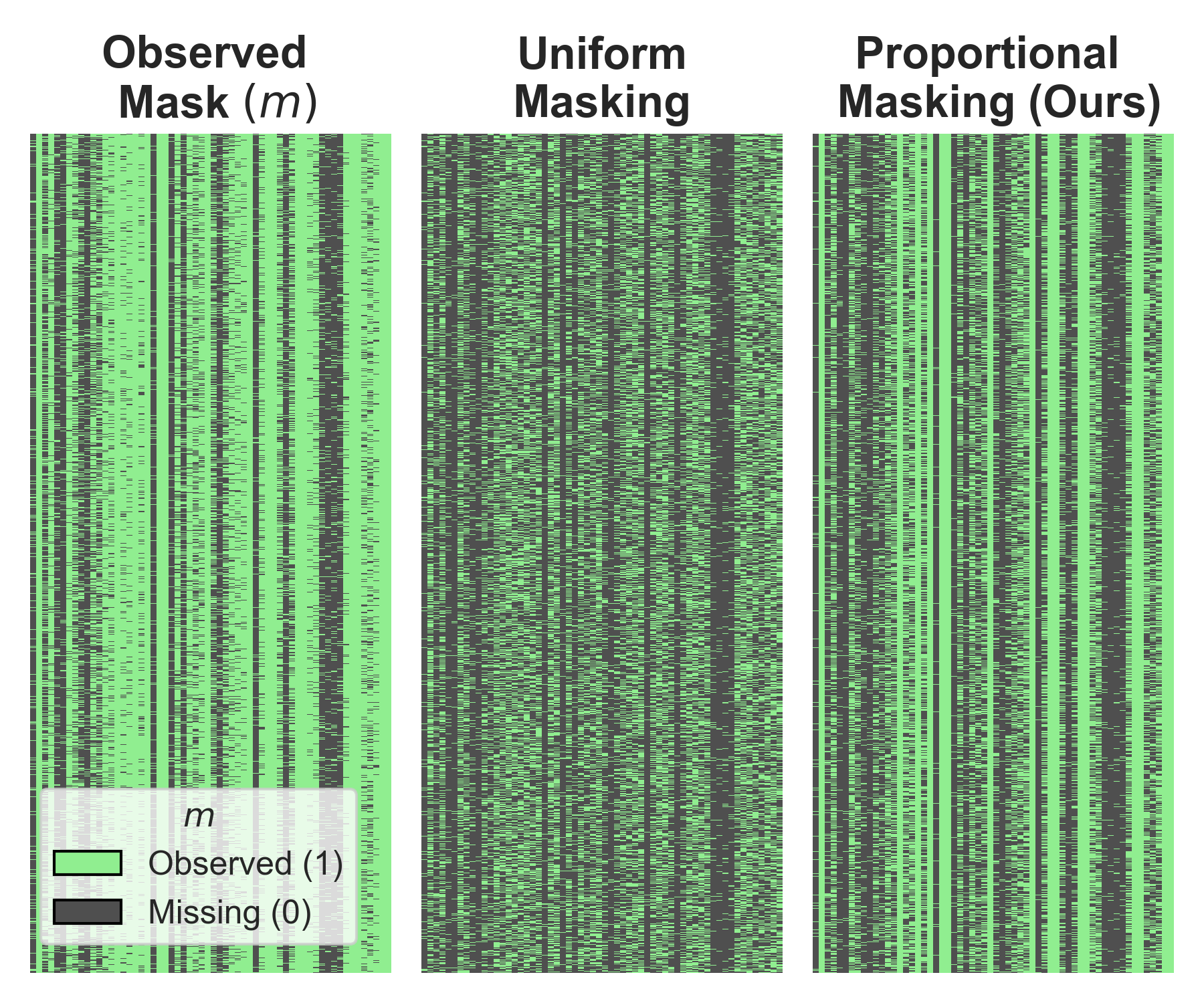}}
    \caption{Masking data inversely to observed proportions, prioritizing \textit{prediction} when data is sparse and \textit{reconstruction} when more data is observed (the last column of (b)).}
    \label{fig:mask_rate}
\end{figure}

\section{Method}
\subsection{Proportional Masking} 
We propose to guide MAEs to emphasize prediction or reconstruction in specific columns using a well-designed masking scheme by leveraging the observed proportions of columns as prior information, defined as:
\begin{equation}
\mathbf{p}_{\mathrm{obs}} := \frac{1}{|B|} \sum_{i\in B} \mathbf{m}_{i} \in \mathbf{R}^d.
\label{observed proportion}
\end{equation}
Since $\delta_{ij} \sim \mathrm{Ber}(\pi_{ij})$, \eqref{observed proportion} serves as an MLE estimate for the column-wise average of the unknown propensity score function, i.e., $p_{\mathrm{obs}, j} \approx \mathbb{E}_{\mathbf{x}}[\pi_{\cdot j}(\mathbf{x};\boldsymbol{\phi})]$.

Our proposed masking function takes the \textit{logit-transformed} value of \eqref{observed proportion}, where we verify the effectiveness of this design choice in Table~\ref{table3}:
\begin{align}
M^{\mathrm{PM}}_{j}(p_{\mathrm{obs}, j};a, b) &:= 
    a\log\left(\frac{1-p_{\mathrm{obs}, j}}{p_{\mathrm{obs}, j}}\right) + b,
\label{eq:logit_mask}
\end{align}
where we clip the output to $[0, 1]$ to prevent undefined behavior as $p_{\mathrm{obs}, j}$ approaches 0 or 1, and $a$ and $b$ are hyperparameters.
Note that this mask is not applied to columns that are fully observed across all rows, i.e., when $p_{\mathrm{obs}, j}=1$.

We aim to achieve two objectives with this design:
(i) to provide neural networks with signals corresponding to the average propensity score function within the batch, which implicitly serve as inductive biases; and
(ii) to emphasize samples inversely proportional to their observed proportions.
Intuitively, $a$ controls the sensitivity to changes in the observed proportion $p_{obs,j}$ (or the logit), and $b$ determines the base masking ratio regardless of $p_{obs,j}$. While the parameters could be learnable, we fix them at $a=0.05$ and $b=0.5$ (behavior shown in Fig.~\ref{fig:mask_rate} (a)), as our grid search across validation datasets indicates these values perform well (Fig.~\ref{grid_search}). 

\subsubsection{Parameter Choice}  We layout the following design principles for selecting parameters in the masking function:
\begin{enumerate}
    \item (Choice of $a$):
    \begin{itemize}
        \item \textit{Sign}. Ensure $a > 0$ to emphasize samples inversely to their observed proportions. 
        \item \textit{Magnitude}. The value of $a$ should be carefully balanced: (i) If $a$ is \textit{too small}, it leads to uniform random masking, and (ii) if $a$ is \textit{too large}, it enforces a hard decision rule where the model \textit{always predicts} when $p_{\mathrm{obs}, j} < b$, and \textit{always reconstructs} otherwise.
    \end{itemize}
     \item (Choice of $b$): $b$ represents a prior weight that determines the baseline importance of prediction. In the absence of prior knowledge, setting $b = 0.5$ is recommended for an equal balance between prediction and reconstruction.
\end{enumerate}


\subsection{Architecture: MLP-Mixers vs. Transformers} 

Transformer-based architectures have been prevalent in many MAE designs in the CV/NLP domains, where stacks of Self-Attention (SA) and feed-forward blocks are utilized for learning intricate pairwise (and higher-order) relationships between different columns. 
Nonetheless, as discussed by~\citet{yan2023t2g}, interactions may exist in multiple groups, and we postulate that this makes the imputation problem harder to solve; learning such relations in a pairwise manner may be inefficient. 
Thus, we propose to utilize the MLP-Mixer architecture \cite{tolstikhin2021mlp}, where token-mixing MLPs with $L^2$ regularization can disconnect combinations of columns through activation functions.
Given the tokenized data (positional information and [CLS] token appended: $\mathbf{x} = q_{\text{tok}}(\tilde{\mathbf{x}} \odot \textbf{m}^+) \in \mathbf{R}^{n\times (d+1)\times c} $), we feed it to encoder-decoder architecture ($h$), where the following basic blocks are stacked in multiple layers:

\begin{itemize}
\item \textbf{Transformer Blocks}
\begin{equation}
\mathbf{x} \leftarrow {{\mathbf{x}}+\text{LN}}({\mathbf{x}} + \text{SA}_{d}(\text{LN}(\mathbf{x}))),
\nonumber
\end{equation}
\begin{equation}
\mathbf{x} \leftarrow {{\mathbf{x}}+\text{LN}}({\mathbf{x}} + \text{MLP}_{c}(\text{LN}(\mathbf{x}))). \label{trf}
\end{equation}
\item \textbf{Mixer Blocks}
\begin{equation}
\mathbf{x} \leftarrow {{\mathbf{x}}+\text{LN}({\mathbf{x}} + \text{MLP}_d(\text{LN}(\mathbf{x})})),
\nonumber
\end{equation}
\begin{equation}
\mathbf{x} \leftarrow {{\mathbf{x}}+\text{LN}}({\mathbf{x}} + \text{MLP}_{c}(\text{LN}(\mathbf{x}))). \label{mix}
\end{equation}
\end{itemize}
The subscript in SA/MLP denotes the dimension in which attention/mixing is applied. Only the token-mixing part (in the $d$ dimension) differs where the MLP replaces SA.


\section{Experiments}

\subsection{Experimental Setup}

\subsubsection{Semi-synthetic Missing Pattern Generation} 
Given a complete dataset, we specify \textit{missing patterns} as follows:
\begin{enumerate}
    \item Monotone Missing
        \begin{itemize}
        \item Generate $\mathcal{M}^{\mathrm{m}}$ with $p^{\mathrm{col}} \in \{0.3, 0.6\}$.
        \item Generate missing entries with a fixed probability of 0.5 (i.e., $P(\delta_{ij} = 1) = 0.5$ for all $j$).
        \end{itemize}
    \item Quasi-Monotone Missing
\begin{itemize}
\item Set $p^{\mathrm{col}} = 0.6$ for $\mathcal{M}_1^{\mathrm{q}}$, and let $\mathcal{M}_2^{\mathrm{q}}:= (\mathcal{M}_1^{\mathrm{q}})^c$.
\item $\forall j \in \mathcal{M}_1^{\mathrm{q}}$, $p_j \sim U(0.95, 0.99)$, $P(\delta_{ij}=1)=p_j$.
\item $\forall j \in \mathcal{M}_2^{\mathrm{q}}$, $p_j \sim U(0.2, 0.8)$, and $P(\delta_{ij}=1)=p_j$.
\end{itemize}
\item General Missing (or Non-Monotone Missing)
\begin{itemize}
\item Set $p^{\mathrm{col}} = 1$ for $\mathcal{M}^{\mathrm{g}}$.
\item $\forall j \in \mathcal{M}^{\mathrm{g}}$, $p_j \sim U(0.2, 0.8)$, and $P(\delta_{ij}=1)=p_j$.
\end{itemize}
\end{enumerate}
Note that $\mathcal{M} := \{ j | P(j\in\mathcal{M}) = p^{\mathrm{col}} \}$ denotes the column indices that will have missing data, with the missing proportion $p^{\mathrm{col}}$. The propensity score function $\pi_{ij}(\mathbf{x}_i; \boldsymbol{\phi}) = P(\delta_{ij} =1|\mathbf{x}_i)$ is specified with the \textit{missing mechanism} and the generated $p_j$. In all settings, we apply the most challenging MNAR. Details are provided in the Appendix.

\subsubsection{Datasets}
We evaluate PMAE along with the baselines using a semi-synthetic setup on nine real-world benchmark datasets~\cite{asuncion2007uci}.

\begin{table}[ht]
\centering
\begin{tabular}{@{}cccc@{}}
\toprule
\textbf{Dataset} & \textbf{\# Samples} & \textbf{\# Categorical} & \textbf{\# Numerical} \\ \midrule
Diabetes & 442 & 1 & 9 \\
Wine & 1,599 & 11 & 0 \\
Obesity & 2,111 & 8 & 8 \\
Bike & 8,760 & 3 & 9 \\
Shoppers & 12,330 & 8 & 10 \\
Letter & 20,000 & 16 & 0 \\
Default & 30,000 & 3 & 9 \\
News & 39,644 & 2 & 46 \\
Adult & 48,842 & 9 & 6 \\ \bottomrule
\end{tabular}
\caption{Benchmark tabular datasets of varying sizes and data types across different domains.}
\label{table1}
\end{table}

\subsection{Baselines and Evaluation}
\subsubsection{Baseline Methods} We compare our model with the following baselines: Naive (numerical:mean and categorical:mode), KNN~ \cite{troyanskaya2001missing} EM~\cite{dempster1977maximum}, MissForest \cite{stekhoven2012missforest},  MiceForest~\cite{shah2014comparison},  MIWAE~\cite{mattei2019miwae}, GAIN~\cite{yoon2018gain}, MIRACLE~\cite{kyono2021miracle}, HyperImpute~\cite{jarrett2022hyperimpute}, TDM~\cite{zhao2023transformed}, and ReMasker~\cite{du2024remasker}.

\setlength{\tabcolsep}{1.0mm}
\begin{table*}[!t]
\centering
\begin{tabular}{@{}ccccccccccccccc@{}}
\toprule
 & \multicolumn{2}{c}{\textbf{Imp. Acc. ($\uparrow$)}} & \multicolumn{2}{c}{${\textbf{R}}^2$ ($\uparrow$)} & \multicolumn{2}{c}{\textbf{Acc. ($\uparrow$)}} & \multicolumn{2}{c}{\textbf{RMSE} ($\downarrow$)} & \multicolumn{2}{c}{\textbf{DT (cls)($\uparrow$)}} & \multicolumn{2}{c}{\textbf{DT (reg)($\uparrow$)}} \\ \midrule
\textbf{Method} & Avg. & Rank & Avg. & Rank & Avg. & Rank & Avg. & Rank & Avg. & Rank & Avg. & Rank & \\ \midrule
Naive & 13.2 ± {\small 3.4} & 12.1 & 0.0 ± {\small0.0} & 12.1 & 36.4 ± {\small8.9} & 11.1 & 27.9 ± {\small4.5} & 10.5 & 88.3 ± {\small2.7} & 8.1 & 23.4 ± {\small1.8} & 7.9  \\
KNN & 38.3 ± {\small6.0}& 4.5 & 26.5 ± {\small7.7}& 5.0 & 54.8 ± {\small7.5}& 4.6 & 20.8 ± {\small2.2}& 5.8 & 88.5 ± {\small2.2}& 6.6 & 24.1 ± {\small2.4}& 7.4 & \\
EM & 36.1 ± {\small5.1}& 5.4 & 26.6 ± {\small6.1}& 4.8 & 49.1 ± {\small6.5}& 6.8 & 20.8 ± {\small2.0}& 5.1 & 89.1 ± {\small2.0}& 6.0 & 24.7 ± {\small2.2}& 5.4 \\
MissForest & 33.7 ± {\small5.0}& 6.6 & 23.4 ± {\small6.0}& 6.2 & 49.5 ± {\small5.8}& 7.3 & 21.0 ± {\small2.3}& 5.7 & 88.5 ± {\small2.1}& 7.1 & 24.5 ± {\small1.3}& 5.9 \\
MiceForest & 33.1 ± {\small6.6}& 7.1 & 20.3 ± {\small8.7}& 8.0 & 53.9 ± {\small7.6}& 5.5 & 23.7 ± {\small2.5}& 9.2 & 87.4 ± {\small2.4}& 9.3 & 22.4 ± {\small3.2}& 9.3 \\
MIWAE & 24.5 ± {\small4.4}& 9.3 & 9.3 ± {\small4.1}& 9.6 & 48.8 ± {\small7.2}& 8.6 & 25.8 ± {\small3.0}& 9.4 & 85.7 ± {\small3.1}& 10.6 & 21.8 ± {\small3.9}& 7.8 \\
GAIN & 17.2 ± {\small3.8}& 11.4 & 4.8 ± {\small4.3}& 11.1 & 39.6 ± {\small7.3}& 10.6 & 29.6 ± {\small2.7}& 10.7 & 88.1 ± {\small1.9}& 7.6 & 23.4 ± {\small2.8}& 8.0 \\
MIRACLE & 26.8 ± {\small7.0}& 8.6 & 14.4 ± {\small7.4}& 8.9 & 42.9 ± {\small8.2}& 8.0 & 29.0 ± {\small3.7}& 10.2 & 88.5 ± {\small2.9}& 7.4 & 21.8 ± {\small1.9}& 9.6 \\
HyperImpute & 36.9 ± {\small6.8}& 5.1 & 26.2 ± {\small9.3}& 5.2 & 53.6 ± {\small8.2}& 5.4 & 21.3 ± {\small2.5}& 6.2 & 88.1 ± {\small2.0}& 6.4 & 23.9 ± {\small1.6}& 6.8 \\
TDM & 22.5 ± {\small4.2}& 9.5 & 7.7 ± {\small4.0}& 9.2 & 45.2 ± {\small8.3}& 9.2 & 22.8 ± {\small2.5}& 7.3 & 88.1 ± {\small2.4}& 8.1 & 24.5 ± {\small2.1}& 6.1 \\
ReMasker & 34.7 ± {\small6.1}& 6.1 & 25.4 ± {\small9.4}& 5.5 & 50.6 ± {\small7.2}& 6.4 & 19.8 ± {\small2.2}& 4.3 & 89.1 ± {\small2.0}& 5.7 & 24.3 ± {\small3.0}& 6.4 \\ \midrule
\textbf{PMAE-trf} & \underline{43.1} ± {\small5.8} & \underline{2.3} & \underline{34.6} ± {\small7.4} & \underline{2.2} & \underline{56.1} ± {\small 8.3} & \underline{3.4} & \underline{18.6} ± {\small2.1} & \underline{2.8} & \textbf{90.1} ± {\small1.3} & \textbf{3.2} & \textbf{25.2} ± {\small2.3} & \textbf{4.5} \\
\textbf{PMAE-mix} & \textbf{44.2} ± {\small5.9} & \textbf{1.8} & \textbf{36.0} ± {\small7.7} & \textbf{1.7} & \textbf{56.4} ± {\small8.4} & \textbf{2.8} & \textbf{18.4} ± {\small1.9} & \textbf{2.5} & \underline{90.1} ± {\small1.2} & \underline{3.6} & \underline{24.8} ± {\small1.8} & \underline{5.2} \\ \bottomrule
\end{tabular}

\caption{Summary result of 13 state-of-the-art methods on 9 datasets, three different \textit{missing patterns}, applied with \textit{MNAR} mechanism, repeated 10 times; \textit{Avg.} are average of 9 (dataset) $\times$ 3 (pattern) $\times$ 10 (seed) = 270 runs, and \textit{Rank} are average of 270 ranked values. For better readability, \textit{Avg.} values are scaled ($\times$100). Details for each datasets can be found in the Appendix.}
\label{table2}
\end{table*}

\begin{figure*}[!ht]
    \centering
        \subfigure[\textbf{Imputation Accuracy (average)}. Every point in each boxplot corresponds to the average value across datasets. Mean value are indicated with ($\blacktriangle$). ]{\includegraphics[width=0.58\textwidth]{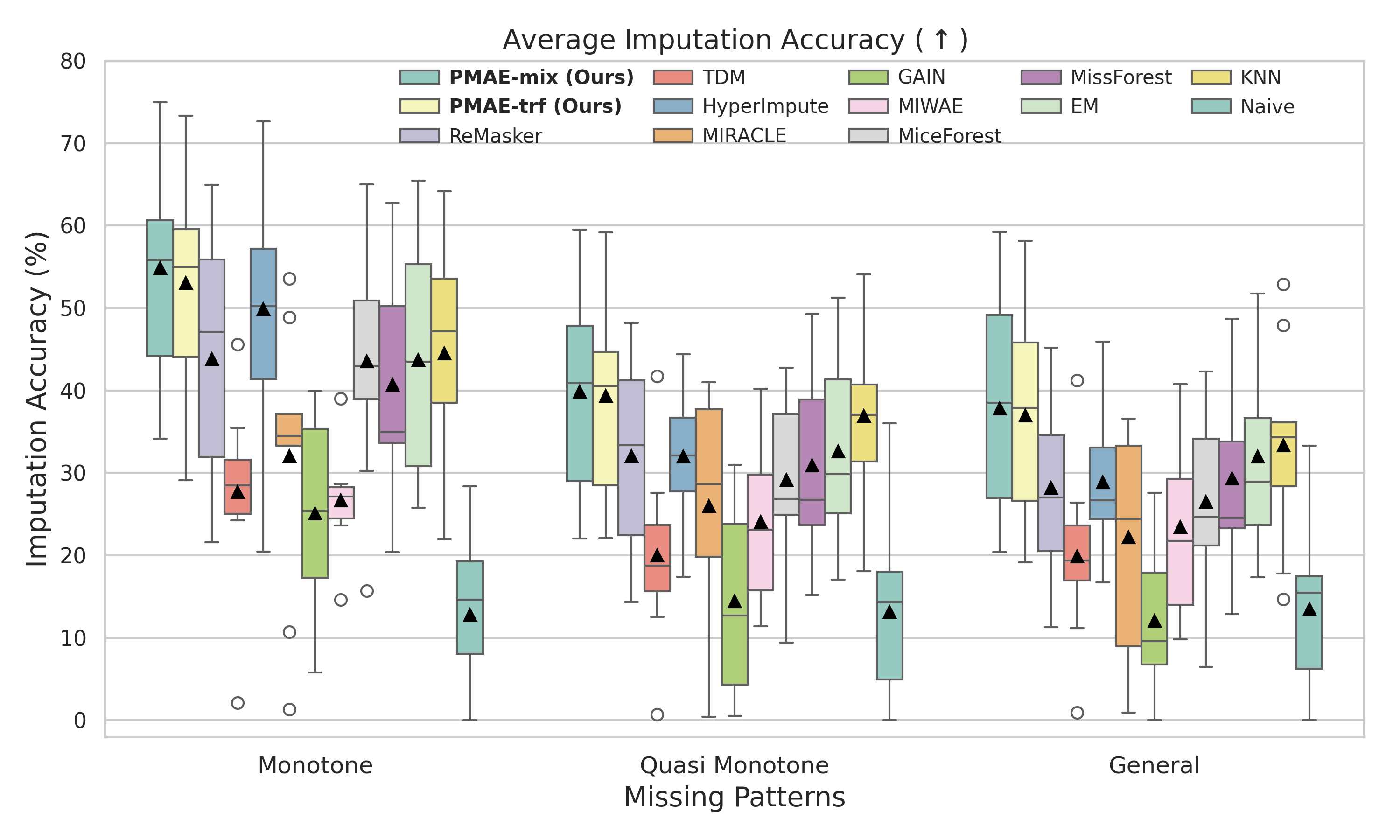}} \hfill%
        \subfigure[\textbf{Imputation Accuracy (rank)}. \textit{PMAE} outperforms all other state-of-the-art methods across all patterns. ]{\includegraphics[width=0.4\textwidth]{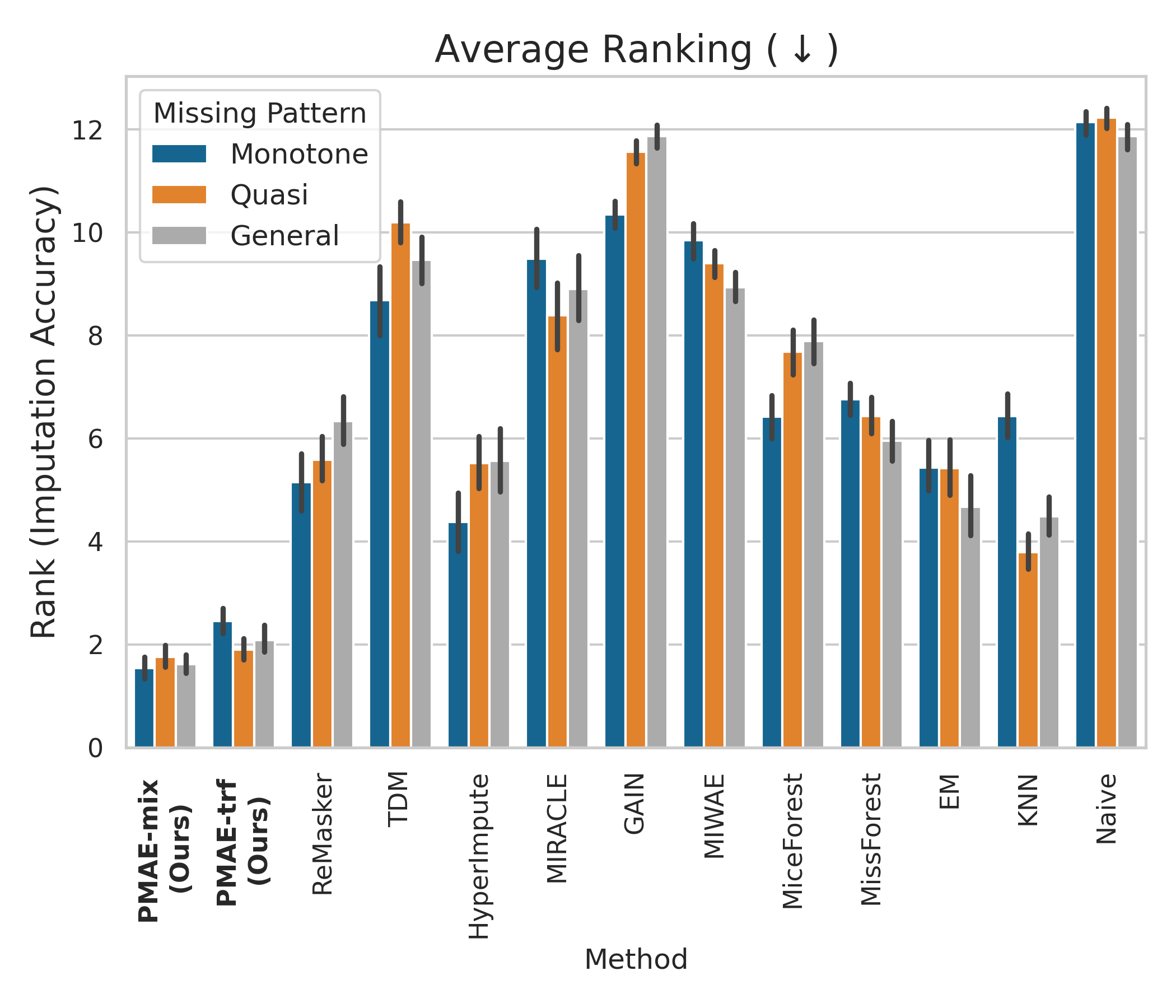}}
    \caption{Imputation accuracy  of state-of-the-art methods across 9 benchmark datasets on missing patterns (\textit{Monotone}, \textit{Quasi Monotone}, and \textit{General}) under NMAR mechanism. Methods are arranged such that the most recent is on the left.}
    \label{fig:IA}
\end{figure*} 

\subsubsection{Parameter Setting} Our implementation mostly follows ReMasker with the same optimization procedures, but introduces (i) a \textit{new loss} formulation in \eqref{main_loss} and (ii) a novel \textit{masking function} $M_j^{\mathrm{PM}}(p_{\mathrm{obs}, j};0.05, 0.5)$, which are applied to (iii) \textit{Transformer} and \textit{MLP-Mixer} architecture.

\subsubsection{Imputation Accuracy} 
Instead of the widely used RMSE, we evaluate imputation performance using a metric we call \textit{Imputation Accuracy}, a weighted average of Accuracy (for categorical columns) and $R^2$ (for numerical columns):
\begin{equation}
\mathrm{Acc}_{j} = \frac{\sum_{i\in I_j^{\mathrm{mis}}} I(\hat{x}_{ij} = x_{ij})}{|I_j^{\mathrm{mis}}|},
\end{equation}

\begin{equation}
R_j^2 = 1 - \frac{\sum_{i\in I_j^{\mathrm{mis}}} (\hat{x}_{ij} - x_{ij})^2}{\sum_{i\in I_j^{\mathrm{mis}}} (\Bar{x}_{ij} - x_{ij})^2},
\end{equation}

\begin{align}
\text{Imp. Acc} := \frac{1}{|D^{\mathrm{mis}}|} \sum_{j \in D^{\mathrm{mis}}} &(\mathrm{Acc}_j \cdot I(j\in D_{c}^{\mathrm{mis}}) \nonumber \\ &+ R_j^2 \cdot I(j\notin D_{c}^{\mathrm{mis}})),
\end{align}
where $D_{c}^{\mathrm{mis}}$ and $I_j^{\mathrm{mis}}$ denote missing categorical column indices and missing row indices for column $j$ respectively. Since the values are adapted to each data type and normalized to 1, this evaluation serves as an intuitive and holistic measure of imputation performance for tabular data.

\subsubsection{Other Measures} While not the primary focus of this study, we also evaluate the following:
\begin{itemize}
    \item \textit{Downstream Task (DT) Performance}: The utility of imputed data measured by evaluating the supervised learning performance; we report the average of test set performances of XGBoost and Linear model (regression, reg: $R^2$, classification, cls: AUROC). 
    \item \textit{RMSE} as in \citet{du2024remasker} for comparison.
\end{itemize}

\subsection{Experimental Results}

\subsubsection{Performance Comparison} 
Table \ref{table2} presents the average values and ranks of imputation methods, while Figure \ref{fig:IA} (a) shows performance variability across datasets/patterns, and Figure \ref{fig:IA} (b) compares the methods' relative performances across patterns.
We summarize key empirical findings:
\begin{itemize}
    \item PMAE-mix achieves the highest imputation accuracy, improving by \textit{at least} 17.5\% (38.3 $\rightarrow$ 44.2).
    \item With our \textit{novel masking function}, we observed an overall improvement of 27.3\% (34.7 $\rightarrow$ 43.1) compared to the ReMasker counterpart overall (at least 7.4\% on \textit{shoppers}, up to 82.3\% on \textit{news} dataset).
    \item Across different missing data patterns, PMAE outperforms ReMasker by 25.1\% on \textit{Monotone} up to 34.1\% for \textit{General} pattern. 
    \item Although the performance of all methods declines as the pattern moves from \textit{Monotone} to \textit{General}, ours maintain consistent relative rankings (as seen in Figure \ref{fig:IA} (b)).
    \item Across data types, we improved $R^2$ by at least 37.6\% (26.6 $\rightarrow$ 36.0), and Accuracy by at least 2.9\% with 11.4\% gain for ReMasker (50.6$\rightarrow$56.4).
    \item With RMSE as the main metric, the relative rank of PMAE are superior and consistent; however, the average rank for different methods shift, placing ReMasker as the second best (6.1 $\rightarrow$ 4.3).  
    \item Our methods (PMAE-trf, PMAE-mix) have superior DT performances over all baselines. Also, Transformer architecture is more suitable than MLP-Mixer architecture for these tasks; perhaps, MLPs, being more flexible, generate overfitted predictions compared to SAs.
\end{itemize}

\subsubsection{Impact of MLPs on Token Mixing}
\begin{figure}[!t]
  \centering
    \includegraphics[width=1.0\linewidth]{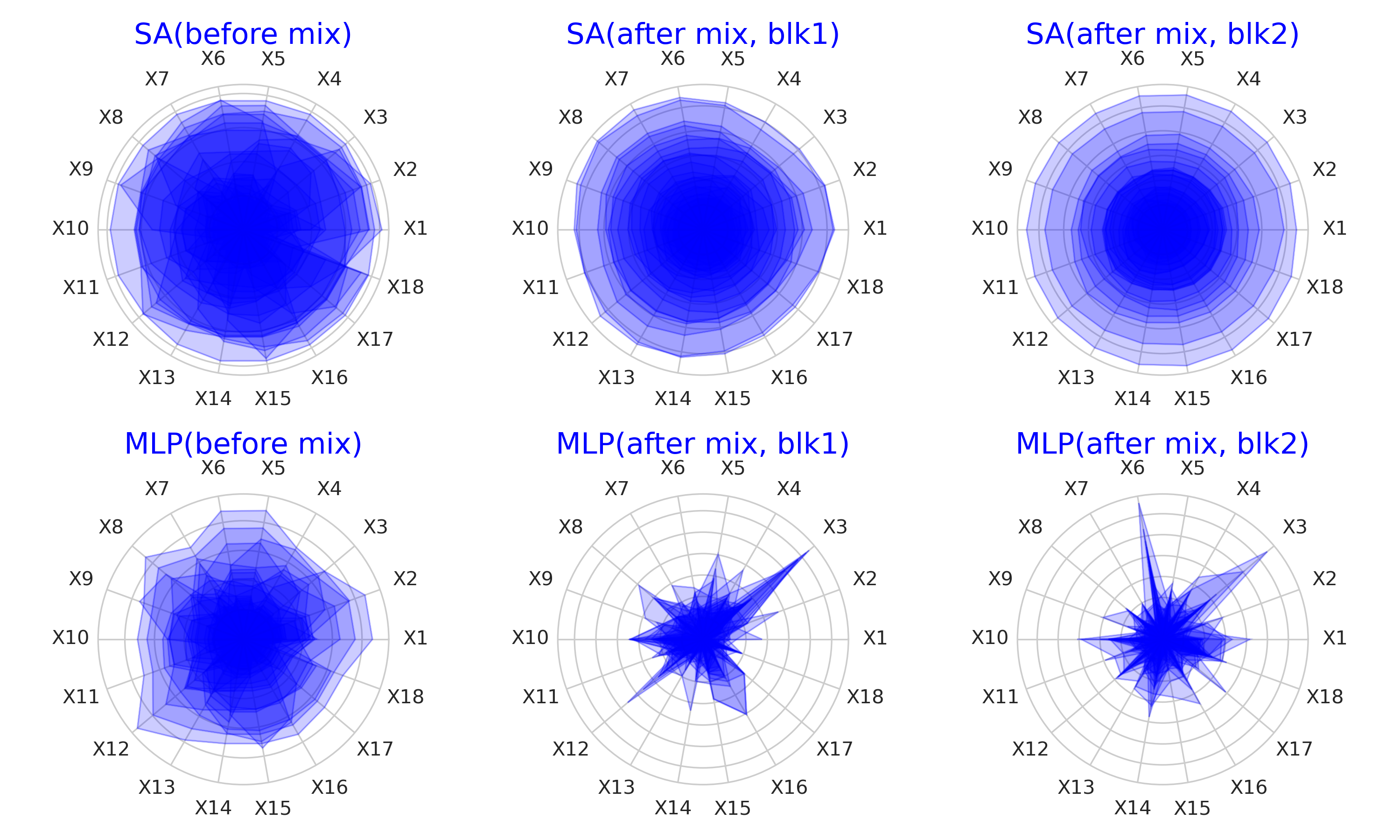}
    \caption{Magnitude comparison of representations after token mixing via. Self-Attention vs. MLP on \textit{Shoppers} dataset. Absolute value is applied for the sake of analysis.}
    \label{token_mix}
\end{figure}

\begin{figure}[!t]
    \centering
        \subfigure[MLP-Mixer]{\includegraphics[width=0.49\linewidth]{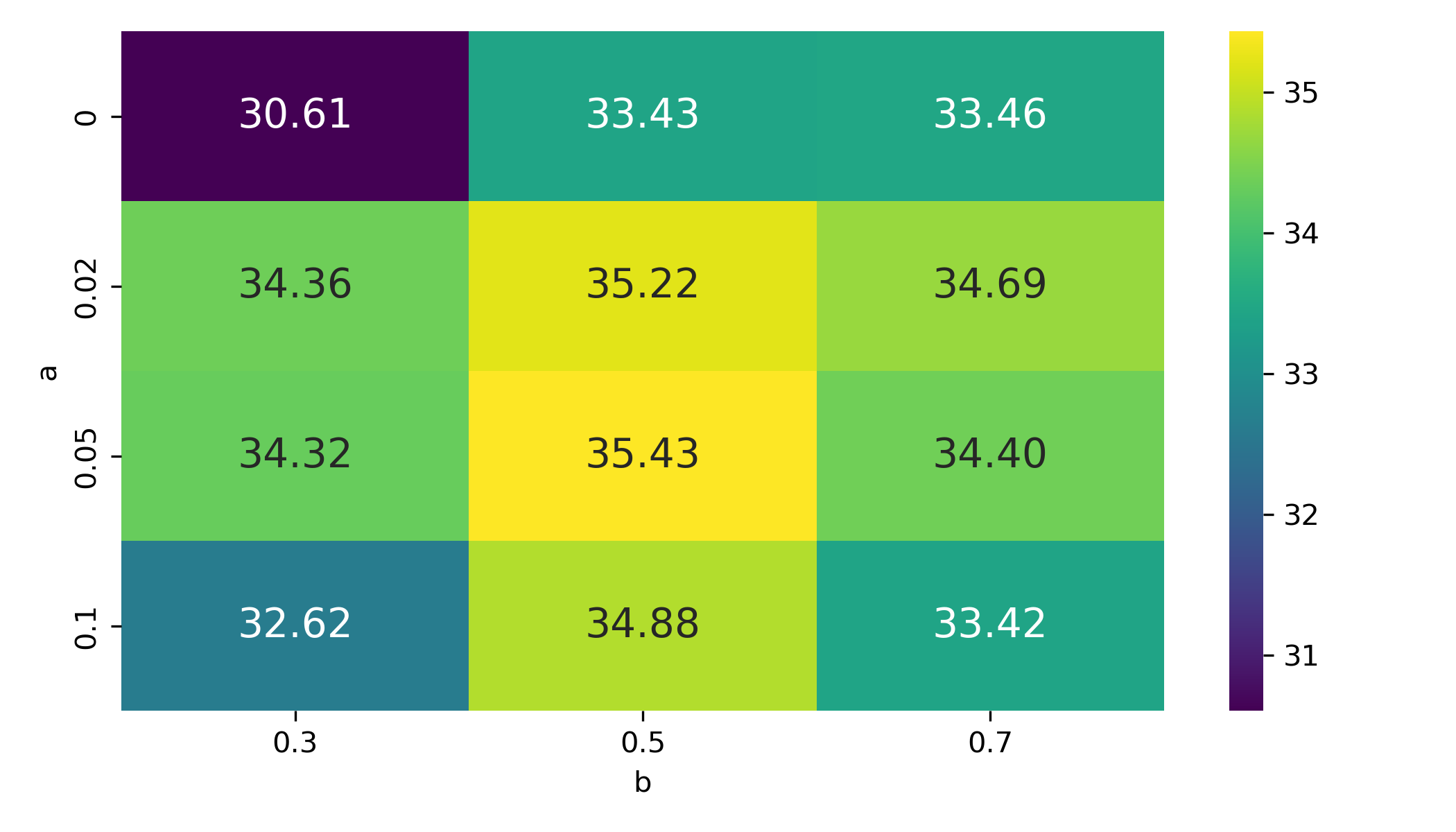}} \hfill%
        \subfigure[Transformer]{\includegraphics[width=0.49\linewidth]{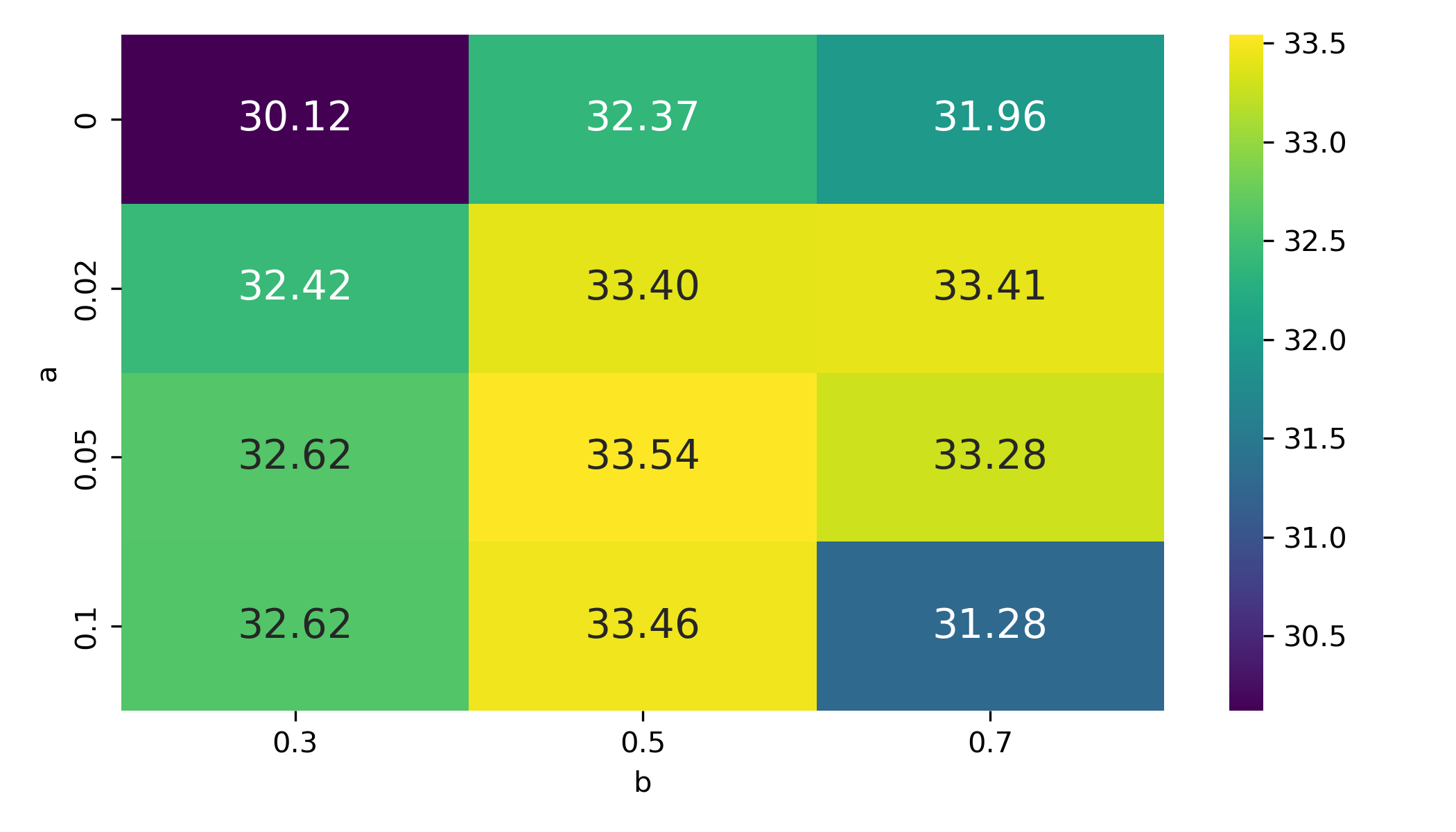}} \hfill%
    \caption{\textbf{Grid search results.} Imputation accuracy across different $a$, $b$ averaged on \textit{Shoppers}, \textit{Wine}, and \textit{Diabetes}.}
    \label{grid_search}
\end{figure}

Figure \ref{token_mix} illustrates the representation vector of SA-based and MLP-based mixing on \textit{shoppers}, highlighting changes in relative magnitudes before and after each mixing block. This shows that MLPs, with activation functions, offer more flexible token mixing compared to SAs, potentially capturing discrete group interactions between columns more effectively.

\subsubsection{Ablation Study: Masking Function Design}
The results in Table 3 support our initial motivations: (i) model performance declines without balancing reconstruction and prediction tasks (\textit{No recon./pred.}); (ii) masking more observed data reduces performance (\textit{Reversed}); (iii) increasing prediction weights for less observed data is beneficial (\textit{Piece-wise}); (iv) the concavity of the masking function is critical as performance improves with a convex function for $p_{\mathrm{obs}, j} \leq 0.5$ and concave function for $p_{\mathrm{obs}, j} > 0.5$, but worsens if concavities are reversed (\textit{Sigmoid-like} vs. \textit{Logit}).

\setlength{\tabcolsep}{0.8 mm}
\begin{table}[!ht]
\centering
\begin{tabular}{@{}c|c|cc@{}}
\toprule
  \multicolumn{1}{l|}{\textbf{Description}} & $M_j(\cdot)$ & \textbf{Perf}. & \textbf{Gain} (\%) \\  \midrule
\multicolumn{1}{c|}{Const.}  &  0.5$\cdot I(p_{\mathrm{obs}, j}<1)$ & 33.4 & 0  \\ 
\multicolumn{1}{c|}{No recon.} &  1.0 &  7.7 &  -76.9  \\
\multicolumn{1}{c|}{No pred.} &  0.0 &  8.6 & -74.3  \\
\multicolumn{1}{c|}{Linear} &    $1 -p_{\mathrm{obs}, j}$ &  34.8& 4.09  \\ 
\multicolumn{1}{c|}{Reversed} &  $p_{\mathrm{obs}, j}$& 31.9& -4.6\\ [+0.7em]
\multicolumn{1}{c|}{Piece-wise} & 
\(\begin{aligned}
  &(1 - p_{\text{obs},j})\cdot I(p_{\text{obs},j} < 0.5)\\ 
  &+ 0.5 \cdot I(p_{\text{obs},j} \geq 0.5)
  \end{aligned}\) & 33.6 & 0.69 \\ [+1.5em] 
\multicolumn{1}{c|}{Sigmoid-like} & $\frac{1}{{\exp(-10(0.5-p_{\mathrm{obs}, j}})+1)}$ & 33.7& 0.78\\ [+1.em] \cmidrule(lr){1-4}  

\multicolumn{1}{c|}{\textbf{Logit (Ours)}} &  $0.05\cdot\log\left(\frac{1-p_{\mathrm{obs}, j}}{ p_{\mathrm{obs}, j}}\right) + 0.5 $&  \textbf{35.4}&  \textbf{5.83}  
\\ \bottomrule
\end{tabular}
\caption{\textbf{Masking function design}. Results of other masking functions averaged on \textit{Shoppers}, \textit{Wine}, and \textit{Diabetes}.}
\label{table3}
\end{table}
\subsubsection{Ablation Study: Loss, $M_j(\cdot)$, Architecture}

\setlength{\tabcolsep}{2 mm}
\begin{table}[!ht]
\centering
\begin{tabular}{@{}cccc@{}}
\toprule
 \textbf{Method} & \textbf{Changes} & \textbf{Perf.} & $\Delta$(\%) \\ \midrule
ReMasker & - & 34.7 & 0.0 \\
ReMasker & Loss: $l^{Re}_{ij} \to$ (\ref{eq.6}) & 36.2 & +4.3 \\
 PMAE-trf & $M_j(\cdot)$: 0.5 $\to$ (\ref{eq:logit_mask}) & 43.1 & +20.0 \\
 PMAE-mix & Architecture: trf $\to$ mix  & 44.2 & +3.0 \\ \bottomrule
\end{tabular}
\caption{Ablations on loss, $M_j(\cdot)$, and architecture.}
\label{table4}
\end{table}

ReMasker uses the following loss in their implementation: $l^{\mathrm{Re}}_{ij} := \frac{(h(\tilde{\mathbf{x}}_{i} \odot \mathbf{m}_{i}^+))_{j}-x_{ij})^2\cdot m_{ij}^+ }{\sum_{i\in B} m_{ij}^+} + \frac{((h(\tilde{\mathbf{x}}_{i} \odot \mathbf{m}_{i}^+))_{j}-x_{ij})^2\cdot m_{ij}^- }{\sum_{i\in B} m_{ij}^-} $, which is equivalent to $m_{ij}^+l_{ij}^{0} + m_{ij}^-l_{ij}^+$ after multiplying by some constant. Then, unlike our formulation in (\ref{eq.6}), \textit{prediction loss} $l_{ij}^{0}$ is weighted by the unmasked parts $m_{ij}^+$. Moreover, ReMasker applies $M(\cdot) =0.5, \forall_{p_{\mathrm{obs}, j}}$. Table~\ref{table4} shows performance gains from adjusting the main loss, masking function, and encoder/decoder blocks.


\section{Conclusion}
Tabular data is inherently heterogeneous, with each column having distinct characteristics and often exhibiting complex patterns of missing values. To address this challenge, we propose PMAE, a simple yet effective strategy that employs a logit-based masking function. This method preserves the distribution of missingness while prioritizing data inversely to their observed proportions. When tested across diverse set of missing patterns, PMAE demonstrated robust performance, consistently surpassing state-of-the-art methods

\subsubsection{Limitations and Future Work}
This study does not examine the relationship between a dataset's covariance structure and the proportional masking scheme, which could provide deeper understanding and broader applicability. Capturing covariance in the presence of missing data, however, remains a challenge. Addressing these issues may enable application of PMAE to high-dimensional tasks like image inpainting.


\section{Acknowledgments}
This work was supported by the National Research Foundation of Korea (NRF) grant funded by the
Korea government (MSIT) (2020R1A2C1A01005949, 2022R1A4A1033384, RS-2023-00217705,
RS-2024-00341749), the MSIT(Ministry of Science and ICT), Korea, under the ICAN(ICT Challenge and Advanced Network of HRD) support program (RS-2023-00259934) supervised by the
IITP(Institute for Information \& Communications Technology Planning \& Evaluation), and the Yonsei
University Research Fund (2024-22-0148).


\bigskip

\bibliography{aaai25}

\clearpage  
\onecolumn
\appendix
\setcounter{secnumdepth}{2}
\numberwithin{table}{section}
\numberwithin{figure}{section}
\numberwithin{equation}{section}
\numberwithin{algorithm}{section}

\section*{Supplementary Material}

The Supplementary Material section is organized as follows.

\begin{enumerate}[A.]
\item \textbf{Implementation Details}: We provide details on the implementation of PMAE and other baseline imputation methods.

\item 
\textbf{Evaluation Procedures}: We provide detailed evaluation procedures to ensure reproducibility, consisting of the following steps: 1) pre-processing step, 2) synthetic generation of missing mechanism and patterns, and 3) evaluation.

\item \textbf{Detailed Results of Table 2}: We present the detailed results of Table 2, covering the evaluated results for each dataset and missing patterns.

\item \textbf{Additional Results}: We present the imputation accuracy across different missing mechanisms (MCAR/MAR/MNAR).

\end{enumerate}

\section{Implementation Details}
\subsection{MAE}
The MAE implementation closely follows \textbf{ReMasker}~\citep{du2024remasker}, referencing their code available at \texttt{https://github.com/tydusky/remasker}, with the following details:
\begin{itemize}
    \item Regarding the optimization procedures (learning rate, learning rate scheduler, and training epochs), we follow ReMasker's configuration. We use different batch sizes depending on the size of the data set for better optimization: \begin{itemize}
        \item For $n < 1,000$: batch size = 128
        \item For $1,000 \leq n < 2,500$: batch size = 256
        \item For $2,500 \leq n < 5,000$: batch size = 512
        \item For $5,000 \leq n < 10,000$: batch size = 1,024
        \item For $10,000 \leq n < 20,000$: batch size = 2,048
        \item For $n \geq 20,000$: batch size = 4,096
    \end{itemize}
    \item Regarding the loss function, we use \eqref{main_loss} for \textbf{PMAE}, and $l_{ij}^{\mathrm{Re}}$ for \textbf{ReMasker} as discussed in the Ablation Study.
    \item Regarding the architecture, we fix the configurations across all datasets:
    \begin{itemize}
        \item We fix the encoder width to 32.
        \item We fix the encoder depth at 6, and the decoder depth at 4.
        \item We fix the attention heads to 4.
        \item For Transformers, we use the same architecture as in ReMasker (stacking multiple layers of \eqref{trf}). 
        \item For MLP-Mixers, we replace SA blocks with \texttt{timm.layers.Mlp} blocks (as in \eqref{mix}), each with \textit{dropout} 0.1, GELU activation and expander width ratio 4.
    \end{itemize}
    \item Regarding the masking function, we used $M_j^{\mathrm{PM}}(p_{\mathrm{obs}, j};0.05, 0.5)$ for \textbf{PMAE}; for \textbf{ReMasker}, we followed the original implementation and set $M_j$ to 0.5 for all columns regardless of the varying levels of observed proportions.
\end{itemize}

\subsection{Baselines}
We also included the results for \textbf{IGRM}~\cite{zhong2023data} in the appendix (although not included in the main paper as the Out-Of-Memory(-) error issue occurred when the dataset size is larger than 8,000).
\\
We provide details on how each of the methods in Table 2 is implemented.
\begin{itemize}
    \item For \textbf{Naive}~\cite{hawthorne2005imputing}, we impute with mean for numerical variables, and mode for categorical variables.
    \item For \textbf{KNN} ~\cite{troyanskaya2001missing}, we use \texttt{KNNImputer} from python library \texttt{Scikit-learn}, with \textit{n\_neighbors} = 5.
    \item We use python library \texttt{Hyperimpute} for the following algorithms, with the following settings:
    \begin{itemize}
        \item \textbf{EM} \cite{dempster1977maximum}:  \textit{max\_it} = 25, \textit{convergence\_threshold} = 1e-6.
        \item \textbf{MissForest} \cite{stekhoven2012missforest}: default setting.
        \item \textbf{GAIN} ~\cite{yoon2018gain}: default setting.
        \item \textbf{MIWAE}~\cite{mattei2019miwae}:  \textit{n\_hidden} = 32, \textit{latent\_size} = 16.
        \item \textbf{MIRACLE}~\cite{kyono2021miracle}: default setting.
        \item \textbf{HyperImpute}: ~\cite{jarrett2022hyperimpute}: \textit{optimizer} = \textit{hyperband}, \textit{classifier\_seed} = [\textit{'logistic\_regression', 'catboost', 'xgboost', 'random\_forest'}], \textit{regression\_ssed} = [\textit{'linear\_regression', 'catboost\_regressor', xgboost\_regressor', 'random_forest\_regressor'}]. 
    \end{itemize}
    \item For \textbf{MiceForest} ~\cite{shah2014comparison}, we use the Python library \texttt{MiceForest}, which uses LightGBM for the replacement of Random Forests, with the aggregate of 5 imputation output.
    \item For \textbf{TDM} ~\cite{zhao2023transformed}, we use the default settings implemented by the authors: \texttt{https://github.com/hezgit/TDM}.
    \item For \textbf{IGRM}~\cite{zhong2023data}, we use the default settings implemented by the authors: \texttt{https://github.com/G-AILab/IGRM} with \textit{max\_epoch} = 8,000 (as the loss curve reaches plateau after this).

\end{itemize}

\section{Detailed Evaluation Procedures}
\subsection{Pre-processing}
\subsubsection{Dataset}
We layout the procedures for processing discrete categorical variables and continuous numerical variables. 
\begin{itemize}
    \item \textbf{Categorical Variables}: To address the varying levels of cardinality and the long-tail distribution of categorical variables, we regroup minor categories into a single common category as follows: 
\begin{itemize}
    \item \textit{Regrouping criteria}: If a category's frequency is less than the following, it is assigned to a common category:
    \begin{enumerate}[i.]
        \item if $ n/100 > 30$, regroup if category's frequency $< \mathrm{round}(n/100)$, 
        \item else, regroup if category's frequency $<$ 30.  
    \end{enumerate}
    \item Apply \texttt{LabelEncoder} function from the \texttt{Scikit-learn} python library and organize categorical variables.
\end{itemize}
\item \textbf{Numerical Variables}: Apply \texttt{QuantileTransformer} for numerical variables to make the distributions symmetric.
\end{itemize}

As a result of the above, we have: $\mathbf{X}\in \mathbb{R}^{n\times (d_n + \sum_{j = 1, ..., c_{K}}K_j})$ where $d_n$ denotes the number of numerical columns and $K_j$ denotes the cardinality of the categorical column $c_j$. 
\subsection{Synthetic Generation of Missing Mechanisms}
We refer to \cite{jarrett2022hyperimpute} for simulating different missing mechanisms. We fit a logistic regression model for the propensity score function $\pi_{ij}(\mathbf{x}; \boldsymbol{\phi}_j) = Pr(\delta_{ij} =1|\mathbf{x};\boldsymbol{\phi}_j)$. Here we describe how the parameters $\boldsymbol{\phi}_j = \{\beta_{j'}\}$ are set:
\begin{itemize}
    \item Choose an observed proportion $p_j$ for the given missing column index ($j\in \mathcal{M}$).
    \item $\beta_{j'} \sim N(0,1)$ with $j^' = 1, ..., d_n, d^{c_1}_1, d^{c_1}_2, ..., d^{c_1}_{|c_1|}, d^{c_2}_{1}, ...,  d^{c_K}_{|c_K|}$, which is the reordered indices of numerical and categorical variables; $d_n$ denotes the number of numerical variables and $d^{c_k}_1$ denotes the first category of the $k$th categorical column.
    \item Choose a mechanism and apply the following before fitting the propensity score function: \begin{itemize}
        \item MCAR: Set $\beta_{j^'} = 0$ for all $j'$, 
        \item MAR: Set $\beta_{j^'} = 0$ for $j' \in \mathcal{M}$, 
        \item MNAR: For \textit{Monotone} missing, set $\beta_j = 0$ for $j' \notin \mathcal{M}$; otherwise, do not change anything;
    \end{itemize}
    \item Solve for $\sigma(\beta_0 + \mathbf{X} \boldsymbol{\beta}$) $= p_j$ to set the values for $\beta_0$ where $\sigma(\cdot)$ is a sigmoid function, 
    \item Generate missingness: $\delta_{ij} \sim Ber(\pi_{ij}(\mathbf{x}; \boldsymbol{\phi}_j))$
\end{itemize}

\subsection{Synthetic Generation of Missing Patterns}
We present the simulated results of missing data patterns, where each line is a KDE-plot for the distribution of the observed proportion of columns. As the pattern becomes more general, the observed proportions across samples vary more uniformly.
\begin{figure*}[!ht]
    \centering
        \subfigure[Monotone]{\includegraphics[width=0.3\textwidth]{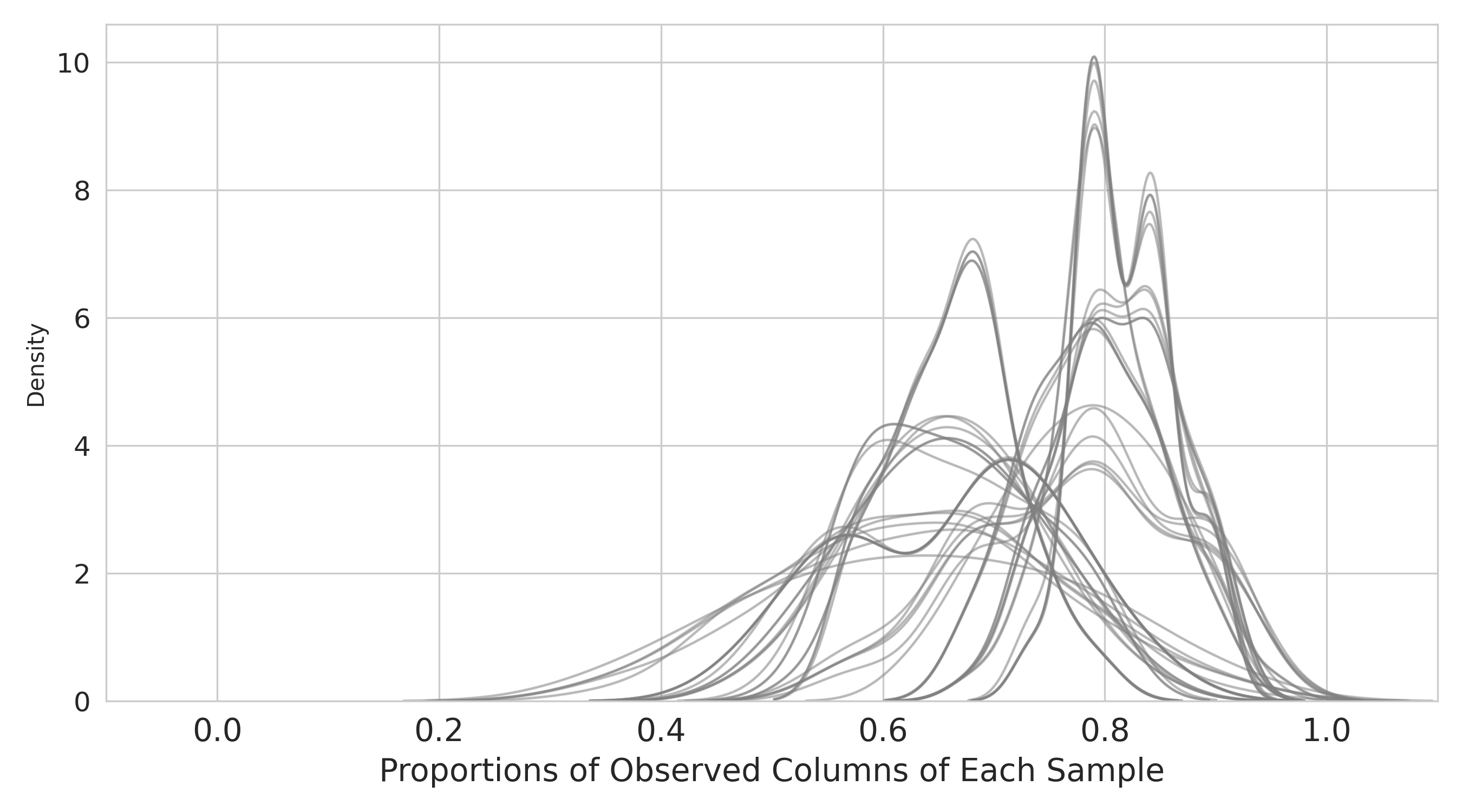}} \hfill%
        \subfigure[Quasi-Monotone]{\includegraphics[width=0.3\textwidth]{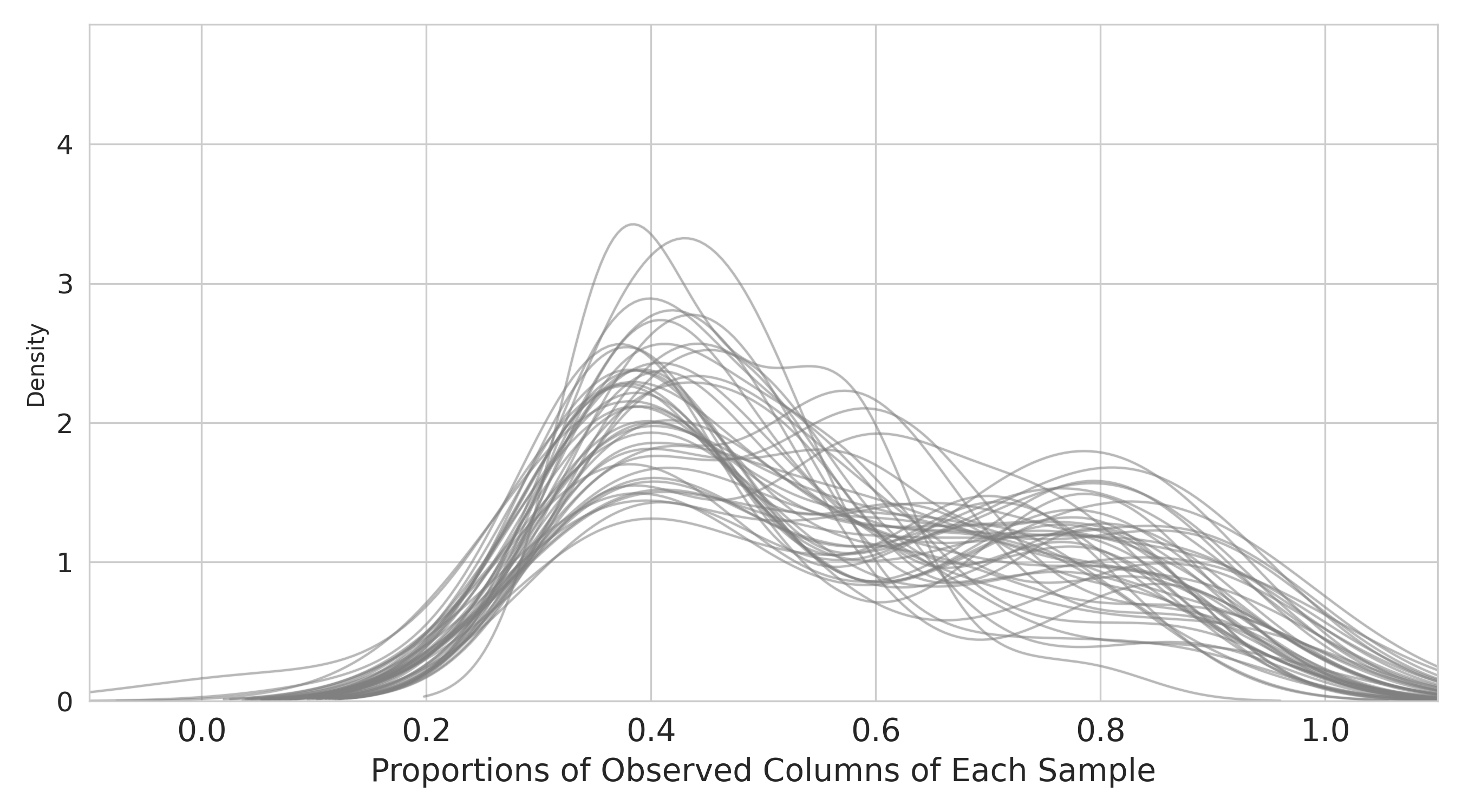}} \hfill%
        \subfigure[General (Non-Monotone)]{\includegraphics[width=0.3\textwidth]{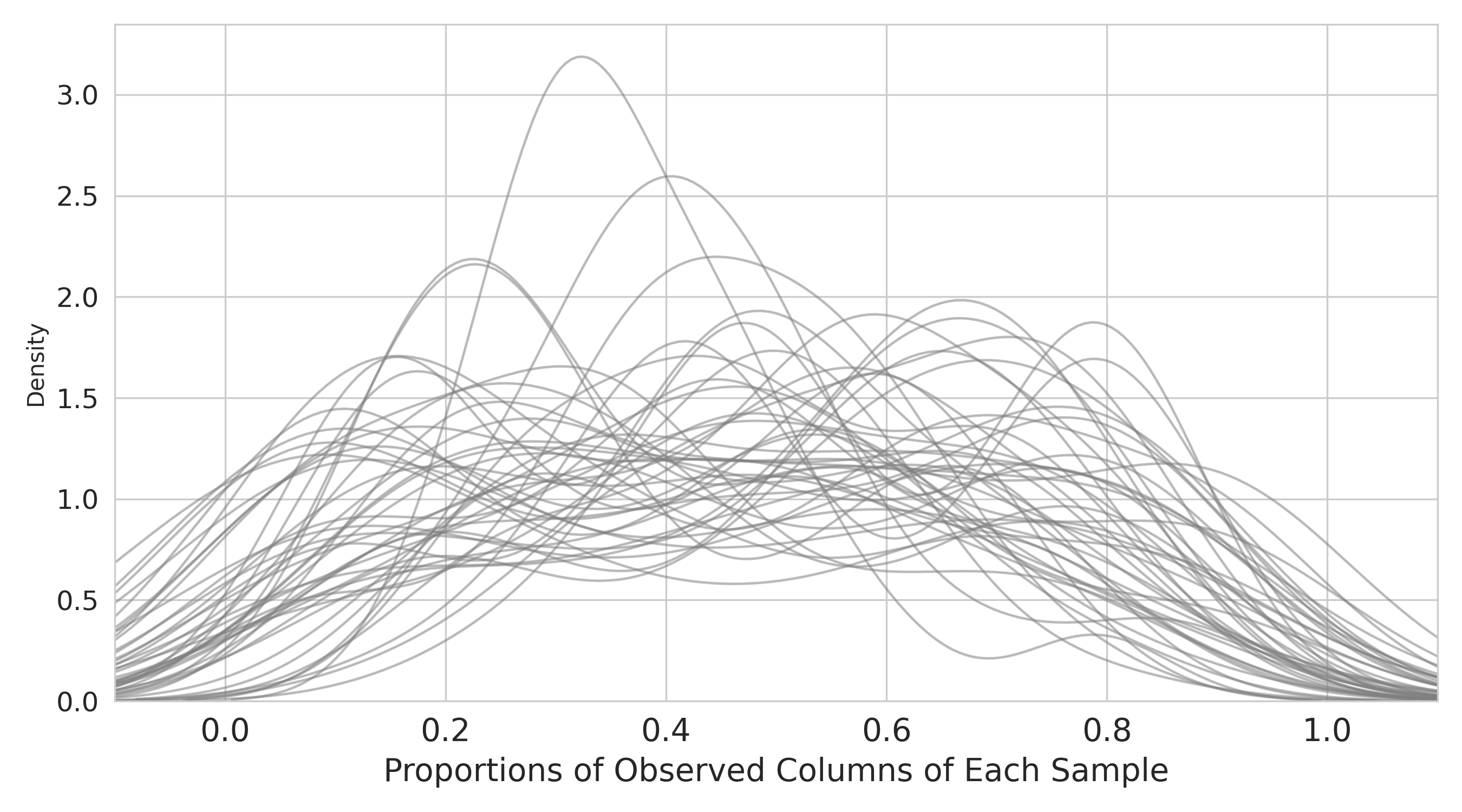}} \hfill%
    \caption{Decrease in proportion of observed column entries moving towards right}
\end{figure*} 
\subsection{Evaluating the Imputation Model}
\subsubsection{Ground Truth and Missing Values} After generating the observed mask, we restore the dimension of the column to $d$. The ground truth data ($\mathbf{X}^*$), incomplete data ($\tilde{\mathbf{X}}$), and imputed data ($\hat{\mathbf{X}}$) are then defined accordingly:
\begin{align}
    \mathbf{X}^* := \mathbf{X}  \odot  (\mathbf{1}_{n\times d} - \mathbf{M})\in \mathbb{R}^{n\times d} \\ 
    \tilde{\mathbf{X}} := \mathbf{X}  \odot  \mathbf{M} + \mathrm{nan}\cdot(\mathbf{1}_{n\times d} - \mathbf{M}) \in \mathbb{R}^{n\times d} \\ 
    \hat{\mathbf{X}} :=  \hat{f}({\tilde{\mathbf X},\mathbf{M}; \hat{\Theta}) }\cdot(\mathbf{1}_{n\times d} - \mathbf{M}) \in \mathbb{R}^{n\times d}
\end{align}

\subsubsection{Evaluation} We feed in ($\tilde{\mathbf{X}}, \mathbf{M}$) to the trained imputation model $\hat{f}({\tilde{\mathbf X},\mathbf{M}; \hat{\Theta})})$ to obtain $\hat{\mathbf{X}}$, and evaluate for ($\hat{\mathbf{X}}$, $\mathbf{X}^*$). Note that for evaluation of categorical columns, we round the imputed values to the nearest integer.

\newpage
\section{Detailed Result of Table 2}
We present individual results from Table 2, summarizing results across datasets and baseline methods under Monotone, Quasi-Monotone, and General (Non-Monotone) missing data patterns, all evaluated with the MNAR mechanism. Each subsection—\textit{Imputation Accuracy}, $R^2$, \textit{Accuracy}, RMSE, RMSE (numerical), RMSE (categorical) and downstream task performance, follows a consistent format.

The results of IGRM~\cite{zhong2023data} are included in this section but are not emphasized in the main paper due to the Out-Of-Memory error on datasets larger than 8,000, which are marked with a '$-$'.

\subsection{Imputation Accuracy}
We present a detailed evaluation of our primary metric, \textit{Imputation Accuracy}, which combines $R^2$ for numerical columns and \textit{Accuracy} for categorical columns.

\begin{table}[!h]
\centering
\resizebox{0.55\textwidth}{!}{
\begin{tabular}{@{}c|cc|ccccccccc@{}}
\toprule
\textbf{Method} & \textbf{Rank} & \textbf{Avg} & \textbf{Diabetes} & \textbf{Wine} & \textbf{Obesity} & \textbf{Bike} & \textbf{Shoppers} & \textbf{Letter} & \textbf{Default} & \textbf{News} & \textbf{Adult} \\ \cmidrule(lr){1-1} \cmidrule(lr){2-3} \cmidrule(lr){4-12} 
Naive & 12.6 & 12.8 & 8.4 & 0.0 & 28.4 & 8.0 & 19.2 & 14.6 & 15.6 & 1.1 & 20.0 \\
KNN & 6.1 & 44.5 & 28.6 & 22.0 & 53.6 & 40.1 & 50.4 & \textbf{56.1} & 64.2 & 47.2 & 38.5 \\
EM & 5.8 & 43.7 & \textbf{43.5} & 25.8 & 49.6 & 32.2 & 55.3 & 30.1 & 65.5 & 60.8 & 30.8 \\
MissForest & 7.2 & 40.7 & 33.6 & 20.4 & 52.0 & 34.9 & 50.2 & 28.3 & 62.8 & 50.2 & 34.3 \\
MiceForest & 6.6 & 43.6 & 30.3 & 15.7 & 50.9 & 39.0 & 50.2 & 40.2 & 65.0 & 57.8 & 43.0 \\
GAIN & 10.8 & 25.1 & 17.3 & 5.8 & 39.9 & 12.5 & 35.3 & 23.1 & 39.6 & 25.4 & 27.1 \\
MIWAE & 10.3 & 29.7 & 23.6 & 14.6 & 39.0 & 27.2 & 35.1 & 27.1 & 32.1 & 40.2  & 28.7 \\
MIRACLE & 9.6 & 32.1 & 1.3 & 10.7 & 37.2 & 33.3 & 35.7 & 34.5 & 48.9 & 53.5 & 33.5 \\
HyperImpute & 4.4 & 49.9 & 36.3 & 20.5 & 51.9 & 50.2 & 57.2 & \underline{46.3} & 72.3 & 72.6 & 41.4 \\
TDM & 8.9 & 27.7 & 35.4 & 24.3 & 45.6 & 25.1 & 31.6 & 31.6 & 25.2 & 2.1 & 28.5 \\
IGRM & 4.1 & 39.9 & 34.5 & \underline{30.1} & 55.2 & - & - & - & - & - & - \\
ReMasker & 6.0 & 43.8 & 31.9 & 21.6 & 48.4 & 47.1 & 58.5 & 24.1 & 64.9 & 55.9 & 42.2 \\ \midrule
\textbf{PMAE-trf} & \underline{2.7} & \underline{53.0} & 36.7 & 29.1 & \underline{57.3} & \underline{55.0} & \underline{59.5} & 44.0 & \underline{73.3} & \underline{72.8} & \underline{49.6} \\
\textbf{PMAE-mix} & \textbf{1.8} & \textbf{54.9} & \underline{41.5} & \textbf{34.2} & \textbf{58.9} & \textbf{55.8} & \textbf{60.6} & 44.2 & \textbf{75.0} & \textbf{74.0} & \textbf{49.8} \\ \bottomrule
\end{tabular}
}
\caption{Imputation Accuracy (Pattern: Monotone Missing, Mechanism: MNAR)}
\end{table}
\vspace{-0.3cm}

\begin{table}[!h]
\centering
\resizebox{0.55\textwidth}{!}{
\begin{tabular}{@{}c|cc|ccccccccc@{}}
\toprule
\textbf{Method} & \textbf{Rank} & \textbf{Avg} & \textbf{Diabetes} & \textbf{Wine} & \textbf{Obesity} & \textbf{Bike} & \textbf{Shoppers} & \textbf{Letter} & \textbf{Default} & \textbf{News} & \textbf{Adult} \\ \cmidrule(lr){1-1} \cmidrule(lr){2-3} \cmidrule(lr){4-12} 
Naive & 12.6 & 13.2 & 4.9 & 0.0 & 36.0 & 7.7 & 18.0 & 14.3 & 14.5 & 0.7 & 22.3 \\
KNN & 3.6 & 36.9 & 22.1 & 18.1 & \textbf{51.7} & 31.4 & 40.7 & \textbf{40.5} & 54.1 & 37.0 & 36.9 \\
EM & 5.8 & 32.6 & 25.9 & 17.1 & 45.3 & 20.9 & 41.3 & 25.1 & 51.3 & 37.0 & 29.9 \\
MissForest & 6.9 & 30.9 & 22.3 & 15.2 & 45.2 & 26.7 & 38.9 & 23.7 & 49.2 & 24.7 & 32.4 \\
MiceForest & 7.8 & 29.2 & 17.1 & 9.4 & 42.7 & 24.9 & 36.1 & 26.8 & 42.1 & 26.2 & 37.2 \\
GAIN & 12.0 & 14.5 & 4.3 & 0.5 & 31.0 & 10.4 & 26.8 & 12.7 & 18.8 & 1.8 & 23.8 \\
MIWAE & 9.7 & 24.1 & 15.8 & 11.4 & 40.2 & 21.1 & 25.3 & 23.1 & 35.7 & 14.2 & 29.8 \\
MIRACLE & 8.2 & 26.0 & 0.4 & 10.3 & 38.0 & 19.9 & 37.7 & 28.7 & 41.0 & 20.8 & 37.2 \\
HyperImpute & 5.6 & 32.0 & 26.2 & 17.4 & 44.4 & 27.7 & 34.3 & \underline{29.7} & 39.4 & 32.1 & 36.7 \\
TDM & 10.6 & 20.0 & 16.2 & 12.5 & 41.8 & 15.6 & 23.3 & 18.8 & 23.7 & 0.7 & 27.6 \\
IGRM & 4.4 & 30.2 & 22.4 & 19.4 & 48.9 & - & - & - & - & - & - \\
ReMasker & 6.2 & 32.1 & 17.1 & 14.4 & 47.1 & 30.1 & 41.2 & 22.4 & 48.2 & 33.4 & 35.0 \\ \midrule
\textbf{PMAE-trf} & \underline{2.2} & \underline{39.4} & \underline{27.2} & \textbf{22.1} & \underline{49.9} & \textbf{38.8} & \underline{43.4} & 28.5 & \underline{59.2} & \underline{44.7} & \underline{40.5} \\
\textbf{PMAE-mix} & \textbf{2.1} & \textbf{39.8} & \textbf{27.7} & \underline{22.0} & 49.1 & \underline{37.8} & \textbf{44.8} & 29.0 & \textbf{59.5} & \textbf{47.9} & \textbf{40.9} \\ \bottomrule
\end{tabular}
}
\caption{Imputation Accuracy (Pattern: Quasi-Monotone Missing, Mechanism: MNAR)}
\end{table}
\vspace{-0.3cm}

\begin{table}[!h]
\centering
\resizebox{0.55\textwidth}{!}{
\begin{tabular}{@{}c|cc|ccccccccc@{}}
\toprule
\textbf{Method} & \textbf{Rank} & \textbf{Avg} & \textbf{Diabetes} & \textbf{Wine} & \textbf{Obesity} & \textbf{Bike} & \textbf{Shoppers} & \textbf{Letter} & \textbf{Default} & \textbf{News} & \textbf{Adult} \\ \cmidrule(lr){1-1} \cmidrule(lr){2-3} \cmidrule(lr){4-12} 
Naive & 12.2 & 13.5 & 6.2 & 0.0 & 33.3 & 7.6 & 17.4 & 15.5 & 16.7 & 0.7 & 24.0 \\
KNN & 4.4 & 33.4 & 17.8 & 14.7 & 47.9 & 28.4 & 34.4 & 33.9 & 52.9 & 36.2 & 34.3 \\
EM & 5.0 & 32.0 & \textbf{23.7} & 17.4 & 45.8 & 23.2 & 36.4 & 24.1 & 51.7 & 36.6 & 29.0 \\
MissForest & 6.4 & 29.3 & 18.1 & 12.8 & 46.0 & 23.8 & 33.8 & 23.3 & 48.7 & 24.5 & 32.9 \\
MiceForest & 8.0 & 26.5 & 14.1 & 6.5 & 42.3 & 21.2 & 29.7 & 24.5 & 41.5 & 24.7 & 34.2 \\
GAIN & 12.3 & 12.1 & 6.8 & 0.0 & 27.6 & 9.6 & 20.8 & 9.1 & 16.0 & 0.9 & 17.9 \\
MIWAE & 9.2 & 23.4 & 14.0 & 9.8 & 40.8 & 20.1 & 22.9 & 21.7 & 38.3 & 14.0 & 29.3 \\
MIRACLE & 9.0 & 22.2 & 1.0 & 8.9 & 36.4 & 18.3 & 32.2 & 24.4 & 36.6 & 9.0 & 33.3 \\
HyperImpute & 5.8 & 28.9 & 22.1 & 16.7 & 45.9 & 26.7 & 24.4 & 25.8 & 33.1 & 31.6 & 33.9 \\
TDM & 9.9 & 19.9 & 16.9 & 11.2 & 41.2 & 17.8 & 23.6 & 19.4 & 21.7 & 1.0 & 26.4 \\
IGRM & 3.5 & 28.9 & 18.5 & 18.3 & \textbf{50.0} & - & - & - & - & - & - \\
ReMasker & 7.1 & 28.2 & 13.5 & 11.3 & 45.2 & 25.4 & 34.6 & 20.5 & 43.1 & 27.0 & 33.6 \\ \midrule
\textbf{PMAE-trf} & \underline{2.3} & \underline{37.0} & 21.5 & \underline{19.2} & \underline{49.5} & \textbf{35.6} & \underline{37.9} & \underline{26.6} & \underline{58.2} & \underline{45.8} & \textbf{38.5} \\
\textbf{PMAE-mix} & \textbf{1.8} & \textbf{37.8} & \underline{23.1} & \textbf{20.4} & 49.2 & \underline{35.2} & \textbf{38.7} & \textbf{27.0} & \textbf{59.2} & \textbf{49.3} & \textbf{38.5} \\ \bottomrule
\end{tabular}
}
\caption{Imputation Accuracy (Pattern: General Missing, Mechanism: MNAR)}
\end{table}
\vspace{-0.3cm}

\newpage

\subsection{Coefficient of determination, $R^2$, (numerical variables only) }
We also show the individual results of $R^2$, which are imputation accuracy results evaluated on numerical variables only. 

\begin{table}[!h]
\centering
\resizebox{0.65\textwidth}{!}{
\begin{tabular}{@{}c|cc|cccccccc@{}}
\toprule
\textbf{Method} & \textbf{Rank} & \textbf{Avg} & \textbf{Diabetes} & \textbf{Wine} & \textbf{Obesity} & \textbf{Bike} & \textbf{Shoppers} & \textbf{Default} & \textbf{News} & \textbf{Adult} \\ \cmidrule(lr){1-1} \cmidrule(lr){2-3} \cmidrule(lr){4-11} 
Naive & 11.5 & 0.0 & 0.0 & 0.0 & 0.0 & 0.0 & 0.0 & 0.0 & 0.0 & 0.0 \\
KNN & 6.6 & 33.5 & 21.2 & 22.0 & 22.0 & 35.6 & 48.7 & 60.8 & 48.6 & 9.3 \\
EM & 5.4 & 37.0 & \textbf{39.2} & 25.8 & 16.5 & 29.5 & 56.3 & 62.1 & 63.4 & 3.5 \\
MissForest & 6.6 & 33.1 & 27.3 & 20.4 & 20.5 & 30.4 & 48.5 & 58.9 & 52.6 & 6.6 \\
MiceForest & 7.1 & 33.2 & 22.7 & 15.7 & 16.9 & 34.0 & 46.0 & 61.2 & 59.1 & 10.2 \\
GAIN & 10.1 & 11.9 & 10.0 & 5.8 & 2.0 & 2.5 & 22.2 & 24.9 & 26.0 & 1.7 \\
MIWAE & 9.6 & 16.4 & 16.0 & 14.6 & 0.9 & 20.4 & 17.5 & 31.2 & 30.5 & 0.0 \\
MIRACLE & 9.2 & 21.5 & 0.0 & 10.7 & 3.0 & 27.5 & 31.3 & 41.7 & 55.3 & 2.2 \\
HyperImpute & 4.7 & 42.5 & 29.9 & 20.5 & 19.1 & 48.7 & 60.5 & 71.1 & 74.9 & 15.3 \\
TDM & 8.4 & 13.5 & 32.3 & 24.3 & 11.4 & 17.9 & 10.2 & 8.7 & 1.3 & 2.2 \\
IGRM & 4.2 & 28.2 & 28.3 & \underline{30.1} & 26.1 & - & - & - & - & - \\
ReMasker & 7.1 & 38.3 & 25.3 & 21.6 & 14.8 & 44.5 & 62.7 & 61.5 & 57.8 & 18.4 \\ \midrule
\textbf{PMAE-trf} & \underline{3.2} & \underline{47.2} & 30.6 & 29.1 & \underline{28.7} & \underline{54.0} & \underline{63.5} & \underline{72.7} & \underline{75.0} & \underline{23.9} \\
\textbf{PMAE-mix} & \textbf{2.3} & \textbf{49.8} & \underline{36.5} & \textbf{34.2} & \textbf{31.9} & \textbf{55.6} & \textbf{64.5} & \textbf{74.8} & \textbf{76.1} & \textbf{24.7} \\ \bottomrule
\end{tabular}
}
\caption{ $R^2$: Imputation Accuracy (numerical only) (Pattern: Monotone Missing, Mechanism: MNAR)}
\end{table}
\vspace{-0.3cm}

\begin{table}[!h]
\centering
\resizebox{0.6\textwidth}{!}{
\begin{tabular}{@{}c|cc|cccccccc@{}}
\toprule
\textbf{Method} & \textbf{Rank} & \textbf{Avg} & \textbf{Diabetes} & \textbf{Wine} & \textbf{Obesity} & \textbf{Bike} & \textbf{Shoppers} & \textbf{Default} & \textbf{News} & \textbf{Adult} \\ \cmidrule(lr){1-1} \cmidrule(lr){2-3} \cmidrule(lr){4-11} 
Naive & 12.7 & 0.0 & 0.0 & 0.0 & 0.0 & 0.0 & 0.0 & 0.0 & 0.0 & 0.0 \\
KNN & 5.2 & 25.2 & 18.8 & 18.1 & \textbf{18.1} & 24.6 & 34.3 & 44.7 & 37.5 & 5.2 \\
EM & 6.0 & 22.3 & \underline{24.4} & 17.1 & 7.9 & 13.8 & 34.1 & 40.9 & 37.7 & 2.3 \\
MissForest & 6.9 & 20.0 & 19.3 & 15.2 & 8.1 & 20.1 & 29.4 & 39.0 & 25.2 & 3.6 \\
MiceForest & 9.0 & 15.6 & 12.8 & 9.4 & 4.4 & 16.6 & 26.0 & 27.9 & 26.1 & 1.9 \\
GAIN & 11.7 & 2.0 & 0.5 & 0.5 & 0.1 & 0.2 & 12.2 & 1.6 & 1.1 & 0.2 \\
MIWAE & 10.1 & 9.6 & 11.1 & 11.4 & 1.6 & 11.7 & 8.5 & 18.6 & 14.1 & 0.0 \\
MIRACLE & 9.2 & 12.8 & 0.0 & 10.3 & 3.4 & 10.1 & 29.2 & 25.3 & 20.5 & 3.5 \\
HyperImpute & 6.3 & 20.3 & 21.6 & 17.4 & 9.3 & 20.1 & 28.1 & 27.9 & 32.2 & 5.9 \\
TDM & 10.3 & 5.0 & 13.7 & 12.5 & 3.6 & 5.0 & 1.9 & 2.8 & 0.1 & 0.3 \\
IGRM & 2.7 & 17.4 & 18.2 & 19.4 & 14.5& - & - & - & - & - \\
ReMasker & 3.5 & 21.7 & 14.4 & 14.4 & 10.6 & 23.3 & 33.5 & 37.6 & 33.7 & 6.0 \\ \midrule
\textbf{PMAE-trf} & \underline{1.7} & \underline{30.0} & 23.8 & \textbf{22.1} & \underline{14.8} & \textbf{34.3} & \underline{37.6} & \underline{51.7} & \underline{45.5} & \textbf{10.0} \\
\textbf{PMAE-mix} & \textbf{1.6} & \textbf{30.5} & \textbf{25.3} & \underline{22.0} & 13.7 & \underline{33.1} & \textbf{39.2} & \textbf{51.9} & \textbf{48.6} & \underline{9.8} \\ \bottomrule
\end{tabular}
}
\caption{ $R^2$: Imputation Accuracy (numerical only) (Pattern: Quasi-Monotone Missing, Mechanism: MNAR)}
\end{table}
\vspace{-0.3cm}

\begin{table}[!h]
\centering
\resizebox{0.65\textwidth}{!}{
\begin{tabular}{@{}c|cc|cccccccc@{}}
\toprule
\textbf{Method} & \textbf{Rank} & \textbf{Avg} & \textbf{Diabetes} & \textbf{Wine} & \textbf{Obesity} & \textbf{Bike} & \textbf{Shoppers} & \textbf{Default} & \textbf{News} & \textbf{Adult} \\ \cmidrule(lr){1-1} \cmidrule(lr){2-3} \cmidrule(lr){4-11} 
Naive & 12.4 & 0.0 & 0.0 & 0.0 & 0.0 & 0.0 & 0.0 & 0.0 & 0.0 & 0.0 \\
KNN & 4.8 & 20.8 & 12.7 & 14.7 & 11.3 & 18.9 & 21.5 & 42.5 & 36.8 & 8.0 \\
EM & 4.5 & 20.6 & \textbf{19.6} & 17.4 & 7.3 & 16.0 & 23.7 & 41.0 & 37.5 & 2.5 \\
MissForest & 6.2 & 16.9 & 12.4 & 12.8 & 7.7 & 15.1 & 18.9 & 38.2 & 25.0 & 5.2 \\
MiceForest & 9.0 & 12.0 & 8.6 & 6.5 & 2.9 & 9.7 & 13.8 & 26.5 & 24.7 & 3.5 \\
GAIN & 12.0 & 0.6 & 0.6 & 0.0 & 0.0 & 0.1 & 4.0 & 0.0 & 0.1 & 0.0 \\
MIWAE & 9.7 & 8.3 & 8.7 & 9.8 & 0.5 & 9.0 & 3.3 & 20.9 & 13.9 & 0.1 \\
MIRACLE & 9.1 & 8.8 & 0.0 & 8.9 & 3.3 & 7.4 & 17.5 & 19.8 & 8.7 & 4.9 \\
HyperImpute & 5.9 & 15.9 & 16.9 & 16.7 & 7.4 & 17.0 & 12.0 & 19.0 & 31.8 & 6.2 \\
TDM & 9.7 & 4.5 & 12.2 & 11.2 & 1.6 & 7.0 & 2.1 & 1.2 & 0.3 & 0.5 \\
IGRM & 3.5 & 15.6 & 13.4 & 18.3 & \textbf{15.1} & - & - & - & - & - \\
ReMasker & 6.3 & 16.1 & 8.7 & 11.3 & 7.2 & 15.8 & 20.5 & 30.8 & 27.3 & 7.2 \\ \midrule
\textbf{PMAE-trf} & \textbf{2.2} & \underline{26.7} & 16.4 & \underline{19.2} & \underline{13.3} & \textbf{28.8} & \underline{25.8} & \underline{49.8} & \underline{46.8} & \textbf{13.9} \\
\textbf{PMAE-mix} & \textbf{1.8} & \textbf{27.7} & \underline{18.2} & \textbf{20.4} & 12.9 & \underline{28.3} & \textbf{26.6} & \textbf{51.0} & \textbf{50.4} & \underline{13.5} \\ \bottomrule
\end{tabular}
}
\caption{ $R^2$: Imputation Accuracy (numerical only) (Pattern: General Missing, Mechanism: MNAR)}
\end{table}
\vspace{-0.3cm}

\newpage
\subsection{Accuracy (categorical variables only)}
We also present the individual results for \textit{Accuracy}, which reflects \textit{Imputation Accuracy} specifically for categorical variables.

\begin{table}[!h]
\centering
\resizebox{0.65\textwidth}{!}{
\begin{tabular}{@{}c|cc|cccccccc@{}}
\toprule
\textbf{Method} & \textbf{Rank} & \textbf{Avg} & \textbf{Diabetes} & \textbf{Obesity} & \textbf{Bike} & \textbf{Shoppers} & \textbf{Letter} & \textbf{Default} & \textbf{News} & \textbf{Adult} \\ \midrule
Naive & 11.6 & 34.6 & 42.6 & 56.8 & 32.2 & 40.2 & 14.6 & 40.3 & 16.9 & 33.3 \\
KNN & 5.4 & 57.8 & 60.1 & 85.2 & 53.6 & 52.6 & \textbf{56.1} & 69.3 & 27.3 & 58.0 \\
EM & 6.9 & 51.7 & 61.8 & 82.7 & 40.3 & 54.6 & 30.1 & 70.5 & 24.4 & 49.1 \\
MissForest & 7.6 & 51.3 & 60.2 & 83.4 & 48.5 & 52.2 & 28.3 & 68.8 & 16.4 & 52.8 \\
MiceForest & 5.0 & 58.9 & 61.9 & 84.9 & 53.9 & 55.3 & 40.2 & 70.8 & 39.1 & 64.8 \\
GAIN & 10.0 & 45.3 & 45.6 & 78.0 & 42.6 & 49.8 & 23.1 & 62.9 & 16.7 & 44.0 \\
MIWAE & 9.6 & 42.7 & 56.5 & 77.1 & 47.6 & 44.5 & 27.1 & 60.8 & 22.5 & 47.8 \\
MIRACLE & 9.9 & 43.5 & 7.6 & 71.3 & 50.8 & 40.8 & 34.5 & 60.1 & 28.9 & 54.4 \\
HyperImpute & 4.5 & 59.7 & \underline{63.1} & 84.7 & 54.9 & 54.2 & \underline{46.3} & 74.2 & 41.3 & 58.8 \\
TDM & 9.1 & 46.9 & 50.2 & 79.7 & 46.5 & 54.9 & 31.6 & 51.5 & 14.4 & 46.1 \\
IGRM & 5.4 & \textbf{72.3} & 60.3 & 84.4 & - & - & - & - & - & - \\
ReMasker & 6.3 & 54.0 & 59.4 & 82.0 & 54.8 & 54.1 & 24.1 & 70.2 & 29.4 & 58.0 \\ \midrule
\textbf{PMAE-trf} & \underline{3.0} & 61.0 & 61.8 & \textbf{85.9} & \textbf{58.1} & \textbf{55.5} & 44.0 & \underline{74.2} & \underline{41.5} & \textbf{66.7} \\
\textbf{PMAE-mix} & \textbf{2.4} & \underline{61.6} & \textbf{63.3} & \underline{85.8} & \underline{56.5} & \underline{56.6} & 44.2 & \textbf{75.0} & \textbf{44.7} & \underline{66.5} \\ \bottomrule
\end{tabular}
}
\caption{ Accuracy (categorical only), (Pattern: Monotone Missing, Mechanism: MNAR)}
\end{table}
\vspace{-0.3cm}

\begin{table}[!h]
\centering
\resizebox{0.65\textwidth}{!}{
\begin{tabular}{@{}c|cc|cccccccc@{}}
\toprule
 \textbf{Method} & \textbf{Rank} & \textbf{Avg} & \textbf{Diabetes} & \textbf{Obesity} & \textbf{Bike} & \textbf{Shoppers} & \textbf{Letter} & \textbf{Default} & \textbf{News} & \textbf{Adult} \\ \cmidrule(lr){1-1} \cmidrule(lr){2-3} \cmidrule(lr){4-11}
Naive & 11.1 & 37.3 & 53.4 & 70.8 & 30.5 & 41.6 & 14.3 & 34.1 & 16.8 & 36.6 \\
KNN & 4.4 & 53.8 & 54.4 & \textbf{84.4} & 51.7 & 49.4 & \textbf{40.5} & 66.8 & 25.9 & 57.6 \\
EM & 7.5 & 46.6 & 40.7 & 81.7 & 42.9 & 51.0 & 25.1 & 65.3 & 18.6 & 47.8 \\
MissForest & 8.0 & 47.5 & 50.5 & 81.4 & 46.6 & \underline{51.7} & 23.7 & 63.0 & 12.5 & 51.2 \\
MiceForest & 6.1 & 51.9 & 59.5 & 80.1 & 49.8 & 49.6 & 26.8 & 61.4 & 27.5 & 60.3 \\
GAIN & 10.8 & 38.0 & 43.0 & 61.1 & 41.3 & 45.9 & 12.7 & 42.1 & 19.0 & 39.1 \\
MIWAE & 8.8 & 47.9 & 61.6 & 77.6 & 49.2 & 47.4 & 23.1 & 58.9 & 16.5 & 49.0 \\
MIRACLE & 6.7 & 44.0 & 5.0 & 72.0 & 48.3 & 49.3 & 28.7 & 62.2 & 27.0 & 59.2 \\
HyperImpute & 6.4 & 51.6 & \textbf{70.4} & 78.2 & 50.2 & 43.2 & \underline{29.7} & 55.0 & \textbf{29.0} & 56.8 \\
TDM & 9.5 & 43.5 & 41.4 & 78.8 & 47.4 & 51.0 & 18.8 & 51.8 & 13.8 & 45.2 \\
IGRM & 5.7 & \textbf{72.4} & \underline{62.3} & 82.4 & - & - & - & - & - & - \\
ReMasker & 6.1 & 49.1 & 44.7 & 82.6 & 50.4 & 51.3 & 22.4 & 62.4 & 24.7 & 54.0 \\ \midrule
\textbf{PMAE-trf} & \underline{4.1} & \underline{53.8} & 58.9 & \underline{84.0} & \textbf{52.1} & 51.3 & 28.5 & \underline{69.3} & 25.7 & \underline{60.5} \\
\textbf{PMAE-mix} & \textbf{3.2} & 53.4 & 50.9 & 83.5 & \underline{51.8} & \textbf{52.6} & 29.0 & \textbf{69.8} & \underline{28.5} & \textbf{61.2} \\ \bottomrule
\end{tabular}
}
\caption{ Accuracy (categorical only) (Pattern: Quasi- Monotone Missing, Mechanism: MNAR)}
\end{table}
\vspace{-0.3cm}

\begin{table}[!h]
\centering
\resizebox{0.65\textwidth}{!}{
\begin{tabular}{@{}c|cc|cccccccc@{}}
\toprule
\textbf{Method} & \textbf{Rank} & \textbf{Avg} & \textbf{Diabetes} & \textbf{Obesity} & \textbf{Bike} & \textbf{Shoppers} & \textbf{Letter} & \textbf{Default} & \textbf{News} & \textbf{Adult} \\ \midrule
Naive & 11.4 & 37.4 & 53.5 & 64.4 & 28.6 & 41.0 & 15.5 & 40.3 & 14.9 & 41.0 \\
KNN & 4.4 & 52.9 & 56.2 & 82.7 & \textbf{54.4} & 52.1 & \textbf{33.9} & 67.9 & 22.8 & 52.9 \\
EM & 6.6 & 48.9 & 54.0 & 82.2 & 43.4 & 53.9 & 24.1 & 67.3 & 18.7 & 47.7 \\
MissForest & 6.8 & 49.7 & \textbf{60.8} & 82.1 & 47.9 & 54.2 & 23.3 & 63.7 & 13.4 & 52.4 \\
MiceForest & 6.0 & 51.1 & 56.1 & 79.4 & 52.6 & 51.3 & 24.5 & 62.8 & 25.9 & 55.8 \\
GAIN & 11.6 & 35.4 & 53.5 & 53.3 & 36.6 & 44.0 & 9.1 & 38.6 & 17.6 & 30.4 \\
MIWAE & 8.3 & 48.0 & 53.8 & 78.6 & 51.1 & 49.6 & 21.7 & 62.9 & 16.8 & 49.9 \\
MIRACLE & 8.1 & 41.2 & 8.6 & 67.2 & 48.5 & 52.3 & 24.4 & 60.4 & 14.7 & 53.3 \\
HyperImpute & 5.8 & 49.7 & \underline{60.7} & 82.5 & 53.0 & 42.2 & 25.8 & 53.1 & \textbf{26.8} & 53.4 \\
TDM & 9.6 & 45.1 & 52.0 & 78.6 & 47.9 & 52.8 & 19.4 & 50.8 & 14.9 & 44.8 \\
IGRM & 6.1 & \textbf{69.8} & 56.9 & 82.8 & - & - & - & - & - & - \\
ReMasker & 7.4 & 48.8 & 48.7 & 81.1 & 51.7 & 53.8 & 20.5 & 60.6 & 21.8 & 52.2 \\ \midrule
\textbf{PMAE-trf} & \underline{3.3} & 53.7 & 59.9 & \textbf{83.9} & \underline{54.2} & \underline{54.6} & 26.6 & \underline{70.3} & 24.0 & \underline{55.9} \\
\textbf{PMAE-mix} & \textbf{2.8} & \underline{54.2} & 60.2 & \underline{83.6} & 54.0 & \textbf{55.5} & \underline{27.0} & \textbf{71.0} & \underline{26.2} & \textbf{56.2} \\ \bottomrule
\end{tabular}
}
\caption{ Accuracy (categorical only) (Pattern: General Missing, Mechanism: MNAR)}
\end{table}
\vspace{-0.3cm}

\newpage
\subsection{RMSE (numerical only)}
We present the individual results for \textit{RMSE} for numerical variables only. Our method continues to outperform in the prior metric. 

\begin{table}[!h]
\centering
\resizebox{0.72\textwidth}{!}{
\begin{tabular}{@{}c|cc|cccccccc@{}}
\toprule
\textbf{Method} & \textbf{Rank} & \textbf{Avg} & \textbf{Diabetes} & \textbf{Wine} & \textbf{Obesity} & \textbf{Bike} & \textbf{Shoppers} & \textbf{Default} & \textbf{News} & \textbf{Adult} \\ \midrule
Naive & 9.1 & 18.8 & 17.8 & 15.4 & 21.6 & 21.3 & 25.2 & 17.0 & 19.8 & 12.4 \\
KNN & 7.1 & 16.4 & 18.0 & 16.0 & 20.5 & 17.4 & 17.6 & 14.2 & 14.4 & 13.0 \\
EM & 6.3 & 16.9 & \textbf{12.8} & \textbf{13.3} & 20.6 & 23.3 & 18.4 & 14.9 & 16.2 & 15.4 \\
MissForest & 5.5 & 15.7 & 17.4 & 15.1 & 19.9 & 16.7 & 16.5 & 13.4 & 13.2 & 13.2 \\
MiceForest & 9.2 & 17.8 & 19.1 & 17.9 & 23.2 & 18.4 & 19.0 & 14.7 & 13.3 & 16.6 \\
GAIN & 9.6 & 19.5 & 18.6 & 18.6 & 24.5 & 23.9 & 20.0 & 17.4 & 18.6 & 14.6 \\
MIWAE & 10.0 & 21.9 & 18.0 & 16.6 & 23.5 & 28.8 & 32.8 & 17.1 & 19.8 & 18.6 \\
MIRACLE & 10.2 & 22.3 & 27.9 & 15.5 & 24.6 & 25.0 & 26.5 & 15.8 & 19.3 & 23.5 \\
HyperImpute & 6.5 & 15.8 & 18.8 & 16.8 & 20.6 & 15.2 & 16.0 & 12.9 & 10.6 & 15.8 \\
TDM & 7.4 & 18.0 & \underline{15.4} & 14.9 & 21.0 & 20.7 & 23.7 & 16.5 & 19.6 & 12.3 \\
IGRM & 4.4 & 16.8 & 16.7 & \underline{14.7} & 19.0 & - & - & - & - & - \\
ReMasker & 4.7 & 14.9 & 17.6 & 15.4 & 20.0 & 15.5 & 14.2 & 13.4 & 12.2 & 11.3 \\ \midrule
\textbf{PMAE-trf} & \underline{4.1} & \underline{14.3} & 17.9 & 16.5 & \underline{18.9} & \underline{14.1} & \underline{13.9} & \textbf{12.2} & \textbf{10.0} & \underline{10.9} \\
\textbf{PMAE-mix} & \textbf{3.2} & \textbf{14.1} & 17.5 & 16.0 & \textbf{18.5} & \textbf{13.6} & \textbf{13.8} & \textbf{12.2} & \underline{10.2} & \textbf{10.7} \\ \bottomrule
\end{tabular}
}
\caption{ RMSE (numerical only) (Pattern: Monotone Missing, Mechanism: MNAR)}
\end{table}
\vspace{-0.3cm}

\begin{table}[!h]
\centering
\resizebox{0.72\textwidth}{!}{
\begin{tabular}{@{}c|cc|cccccccc@{}}
\toprule
\textbf{Method} & \textbf{Rank} & \textbf{Avg} & \textbf{Diabetes} & \textbf{Wine} & \textbf{Obesity} & \textbf{Bike} & \textbf{Shoppers} & \textbf{Default} & \textbf{News} & \textbf{Adult} \\ \midrule
Naive & 8.8 & 18.9 & 17.1 & 15.7 & 21.1 & 18.8 & 25.6 & 19.2 & 19.9 & 13.8 \\
KNN & 5.7 & 16.5 & 17.2 & 15.5 & \underline{19.3} & 15.8 & 18.9 & 15.2 & 16.2 & 14.1 \\
EM & 5.5 & 17.3 & 16.2 & 13.9 & 20.1 & 19.0 & 21.2 & 16.8 & 17.3 & 14.1 \\
MissForest & 5.1 & 16.5 & 16.9 & 15.4 & 19.8 & 15.9 & 19.1 & 14.9 & 16.9 & 13.4 \\
MiceForest & 10.2 & 19.6 & 18.8 & 17.3 & 22.7 & 19.6 & 23.5 & 18.2 & 19.0 & 17.5 \\
GAIN & 11.8 & 24.8 & 20.8 & 24.3 & 30.0 & 23.8 & 24.1 & 24.3 & 33.6 & 17.3 \\
MIWAE & 10.3 & 22.3 & 17.3 & 15.6 & 23.9 & 25.2 & 33.9 & 20.9 & 21.4 & 20.4 \\
MIRACLE & 10.4 & 23.3 & 27.0 & 18.1 & 24.4 & 24.8 & 28.0 & 20.5 & 24.9 & 19.0 \\
HyperImpute & 6.3 & 17.5 & 16.9 & 15.3 & 19.4 & 16.3 & 21.8 & 18.2 & 17.8 & 14.2 \\
TDM & 7.4 & 18.5 & \underline{16.4} & \underline{14.8} & 20.8 & 18.5 & 25.1 & 19.0 & 19.5 & 13.8 \\
IGRM & 6.1 & 17.4 & 16.9 & 15.0 & 20.4 & - & - & - & - & - \\
ReMasker & 4.6 & 16.1 & 17.6 & 15.3 & 19.6 & 15.4 & \underline{17.7} & 14.4 & 15.6 & 13.3 \\ \midrule
\textbf{PMAE-trf} & \textbf{3.1} & \textbf{15.7} & 17.1 & 15.3 & \textbf{19.1} & \textbf{14.5} & 17.8 & \textbf{14.0} & \underline{14.6} & \textbf{13.0} \\
\textbf{PMAE-mix} & \underline{3.6} & \textbf{15.7} & 17.2 & 15.6 & 19.4 & \underline{14.8} & \textbf{17.5} & \underline{14.1} & \textbf{14.3} & \textbf{13.0} \\ \bottomrule
\end{tabular}
}
\caption{ RMSE (numerical only) (Pattern: Quasi-Monotone Missing, Mechanism: MNAR)}
\end{table}
\vspace{-0.3cm}

\begin{table}[!h]
\centering
\resizebox{0.72\textwidth}{!}{
\begin{tabular}{@{}c|cc|cccccccc@{}}
\toprule
\textbf{Method} & \textbf{Rank} & \textbf{Avg} & \textbf{Diabetes} & \textbf{Wine} & \textbf{Obesity} & \textbf{Bike} & \textbf{Shoppers} & \textbf{Default} & \textbf{News} & \textbf{Adult} \\ \midrule
Naive & 8.3 & 19.4 & 18.8 & 16.7 & 21.8 & 19.6 & 27.0 & 19.1 & 19.9 & 12.7 \\
KNN & 5.7 & 17.7 & 18.5 & 16.7 & 21.2 & 17.4 & 22.7 & 15.4 & 16.8 & 12.8 \\
EM & 5.1 & 18.1 & \textbf{17.3} & \textbf{15.5} & 21.0 & 21.0 & 24.2 & 15.8 & 17.4 & 12.5 \\
MissForest & 5.0 & 17.7 & 18.3 & 16.8 & 21.0 & 17.1 & 23.4 & 15.2 & 17.3 & 12.3 \\
MiceForest & 10.2 & 21.0 & 20.7 & 18.7 & 25.2 & 21.5 & 27.4 & 18.9 & 19.5 & 15.9 \\
GAIN & 12.9 & 34.6 & 34.9 & 33.1 & 35.9 & 33.7 & 32.6 & 35.7 & 41.5 & 29.1 \\
MIWAE & 9.7 & 23.3 & 18.1 & 16.3 & 25.0 & 30.0 & 33.9 & 20.5 & 23.2 & 19.4 \\
MIRACLE & 10.6 & 26.5 & 28.8 & 21.5 & 26.3 & 26.4 & 32.5 & 24.0 & 38.0 & 14.5 \\
HyperImpute & 7.3 & 19.6 & 18.3 & 16.3 & 21.8 & 18.8 & 29.5 & 21.7 & 18.1 & 12.6 \\
TDM & 6.9 & 19.1 & \underline{17.8} & 15.9 & 21.8 & 19.3 & 26.5 & 19.1 & 19.5 & 12.8 \\
IGRM & 6.3 & 18.6 & 18.9 & \underline{15.7} & 21.3 & - & - & - & - & - \\
ReMasker & 5.6 & 17.9 & 19.6 & 16.5 & 21.5 & 17.6 & 22.4 & 16.4 & 17.2 & 12.2 \\ \midrule
\textbf{PMAE-trf} & \underline{3.0} & \underline{16.7} & 18.3 & 16.4 & \underline{20.6} & \textbf{16.1} & \underline{21.2} & \textbf{14.1} & \underline{15.1} & \textbf{11.9} \\
\textbf{PMAE-mix} & \textbf{2.8} & \textbf{16.6} & 18.2 & 16.4 & \textbf{20.4} & \textbf{16.1} & \textbf{21.0} & \textbf{14.1} & \textbf{14.9} & \textbf{11.9} \\ \bottomrule
\end{tabular}
}
\caption{ RMSE (numerical only) (Pattern: General Missing, Mechanism: MNAR)}
\end{table}
\vspace{-0.3cm}

\newpage
\subsection{RMSE (categorical)}
We present the individual results for \textit{RMSE} for categorical variables only. Our method continues to outperform in the prior metric. 

\begin{table}[!h]
\centering
\resizebox{0.72\textwidth}{!}{
\begin{tabular}{@{}c|cc|cccccccc@{}}
\toprule
\textbf{Method} & \textbf{Rank} & \textbf{Avg} & \textbf{Diabetes} & \textbf{Obesity} & \textbf{Bike} & \textbf{Shoppers} & \textbf{Letter} & \textbf{Default} & \textbf{News} & \textbf{Adult} \\ \midrule
Naive & 12.2 & 43.9 & 75.4 & 47.0 & 49.7 & 36.9 & 22.6 & 39.6 & 39.7 & 40.4 \\
KNN & 7.0 & 29.5 & 63.0 & 26.9 & 25.6 & 27.9 & 7.6 & 30.1 & 30.3 & 24.6 \\
EM & 5.8 & 28.1 & \textbf{49.1} & 27.6 & 28.8 & 24.9 & 14.0 & 25.5 & 30.1 & 24.6 \\
MissForest & 6.4 & 29.4 & 62.7 & 25.7 & 25.7 & 26.5 & 13.6 & 27.7 & 28.5 & 24.3 \\
MiceForest & 9.3 & 32.0 & 61.6 & 27.5 & 28.9 & 32.6 & 12.3 & 31.4 & 34.1 & 27.6 \\
GAIN & 8.1 & 30.7 & 58.5 & 28.0 & 31.2 & 27.5 & 15.7 & 26.5 & 31.6 & 26.3 \\
MIWAE & 9.2 & 31.3 & 53.8 & 34.6 & 28.4 & 30.9 & 17.2 & 23.8 & 30.9 & 30.8 \\
MIRACLE & 10.2 & 36.0 & 54.4 & 38.6 & 38.0 & 39.5 & 17.3 & 34.3 & 33.9 & 32.2 \\
HyperImpute & 5.5 & 27.8 & 60.5 & 24.3 & 23.9 & 27.0 & 9.0 & 27.6 & \underline{25.6} & 24.6 \\
TDM & 8.3 & 31.4 & 57.5 & 32.5 & 29.4 & 27.0 & 15.8 & 30.1 & 34.0 & 25.3 \\
IGRM & 4.3 & 38.9 & 53.6 & \underline{24.2} & - & - & - & - & - & - \\
ReMasker & 4.4 & 26.8 & 53.5 & 25.4 & 24.0 & \textbf{23.9} & 14.6 & 24.5 & 26.8 & 21.4 \\ \midrule
\textbf{PMAE-trf} & \underline{2.8} & \underline{25.3} & 55.6 & 24.3 & \underline{21.9} & 24.2 & \underline{8.4} & \underline{23.1} & \textbf{24.9} & \underline{19.8} \\
\textbf{PMAE-mix} & \textbf{2.2} & \textbf{24.6} & \underline{50.7} & \textbf{23.5} & \textbf{21.7} & \underline{24.0} & \underline{8.4} & \textbf{23.0} & 26.5 & \textbf{19.5} \\ \bottomrule
\end{tabular}}
\caption{ RMSE (categorical only) (Pattern: Monotone Missing, Mechanism: MNAR)}
\end{table}
\vspace{-0.3cm}

\begin{table}[!h]
\centering
\resizebox{0.72\textwidth}{!}{
\begin{tabular}{@{}c|cc|cccccccc@{}}
\toprule
\textbf{Method} & \textbf{Rank} & \textbf{Avg} & \textbf{Diabetes} & \textbf{Obesity} & \textbf{Bike} & \textbf{Shoppers} & \textbf{Letter} & \textbf{Default} & \textbf{News} & \textbf{Adult} \\ \midrule
Naive & 11.3 & 42.4 & 63.5 & 38.8 & 52.9 & 39.0 & 23.4 & 42.4 & 40.7 & 38.8 \\
KNN & 6.5 & 31.1 & 62.1 & 28.0 & 26.3 & 32.6 & 10.2 & 29.9 & 31.8 & 28.0 \\
EM & 5.2 & 29.4 & 50.0 & 29.0 & 27.4 & 28.9 & 15.4 & 25.8 & 31.7 & 27.4 \\
MissForest & 6.7 & 31.9 & 64.5 & 28.9 & 27.0 & 30.5 & 15.2 & 28.7 & 31.5 & 28.7 \\
MiceForest & 9.6 & 35.1 & 58.9 & 30.4 & 32.8 & 38.4 & 15.8 & 31.9 & 39.2 & 33.1 \\
GAIN & 10.6 & 39.9 & 67.9 & 36.7 & 34.0 & 33.6 & 24.8 & 35.9 & 52.7 & 33.8 \\
MIWAE & 9.3 & 35.1 & 52.3 & 38.3 & 28.0 & 42.3 & 18.9 & 33.9 & 34.3 & 32.8 \\
MIRACLE & 10.2 & 41.3 & 66.5 & 40.9 & 42.7 & 45.0 & 27.9 & 34.5 & 39.2 & 33.7 \\
HyperImpute & 6.7 & 30.2 & \textbf{46.7} & 30.2 & 29.1 & 30.7 & 13.2 & 30.3 & 32.5 & 28.5 \\
TDM & 8.0 & 33.9 & 63.3 & 33.4 & 32.1 & 30.9 & 18.1 & 30.1 & 33.9 & 28.9 \\
IGRM & 5.0 & 38.9 & 51.4 & 26.3 & - & - & - & - & - & - \\
ReMasker & 4.0 & 28.6 & 56.7 & 25.0 & 25.9 & \underline{28.1} & 14.4 & 23.9 & 31.0 & 24.0 \\ \midrule
\textbf{PMAE-trf} & \underline{2.3} & \underline{26.8} & 51.4 & \textbf{24.0} & \textbf{23.8} & 28.3 & 12.3 & \textbf{23.2} & \underline{28.4} & \textbf{23.0} \\
\textbf{PMAE-mix} & \textbf{2.0} & \textbf{26.6} & \underline{49.7} & \underline{24.5} & \underline{24.2} & \textbf{28.0} & \underline{12.2} & \textbf{23.2} & \textbf{28.1} & \textbf{23.0} \\ \bottomrule
\end{tabular}
}
\caption{ RMSE (categorical only) (Pattern: Quasi-Monotone Missing, Mechanism: MNAR)}
\end{table}
\vspace{-0.3cm}

\begin{table}[!h]
\centering
\resizebox{0.72\textwidth}{!}{
\begin{tabular}{@{}c|cc|cccccccc@{}}
\toprule
\textbf{Method} & \textbf{Rank} & \textbf{Avg} & \textbf{Diabetes} & \textbf{Obesity} & \textbf{Bike} & \textbf{Shoppers} & \textbf{Letter} & \textbf{Default} & \textbf{News} & \textbf{Adult} \\ \midrule
Naive & 11.5 & 43.4 & 67.3 & 40.9 & 53.3 & 44.9 & 22.6 & 37.4 & 40.4 & 40.5 \\
KNN & 6.3 & 32.7 & 65.9 & 30.0 & 27.7 & 34.4 & 13.6 & 26.9 & 32.9 & 30.3 \\
EM & 4.7 & 29.9 & \underline{48.8} & 30.5 & 29.2 & 30.3 & 16.2 & 23.6 & 31.8 & 28.4 \\
MissForest & 6.1 & 32.0 & 61.7 & 29.6 & 28.6 & 32.1 & 16.1 & 25.8 & 31.7 & 30.7 \\
MiceForest & 9.8 & 37.1 & 66.1 & 32.5 & 34.3 & 39.8 & 18.2 & 29.9 & 41.0 & 35.2 \\
GAIN & 11.7 & 44.6 & 64.2 & 45.1 & 48.3 & 36.2 & 31.7 & 37.6 & 50.1 & 43.3 \\
MIWAE & 9.1 & 35.8 & 55.2 & 39.1 & 29.5 & 42.1 & 20.2 & 31.7 & 35.9 & 32.6 \\
MIRACLE & 9.8 & 41.8 & 59.3 & 41.7 & 42.8 & 46.9 & 28.4 & 33.4 & 48.7 & 33.1 \\
HyperImpute & 7.2 & 33.2 & 62.1 & 29.2 & 30.0 & 33.2 & 16.4 & 29.1 & 32.7 & 33.0 \\
TDM & 7.5 & 33.3 & 55.7 & 37.4 & 30.2 & 32.7 & 18.2 & 28.4 & 33.9 & 29.6 \\
IGRM & 5.0 & 40.8 & 54.9 & 26.7 & - & - & - & - & - & - \\
ReMasker & 4.7 & 30.2 & 55.4 & 27.4 & 28.5 & 29.9 & 17.8 & 23.6 & 32.5 & 26.6 \\ \midrule
\textbf{PMAE-trf} & \underline{1.9} & \underline{27.5} & 49.1 & \textbf{25.8} & \textbf{24.9} & \underline{29.6} & \underline{14.2} & \underline{21.7} & \underline{29.5} & \textbf{24.9} \\
\textbf{PMAE-mix} & \textbf{1.8} & \textbf{27.3} & \textbf{48.3} & \underline{26.0} & \underline{25.2} & \textbf{29.3} & \underline{14.2} & \textbf{21.6} & \textbf{29.2} & \textbf{24.9} \\ \bottomrule
\end{tabular}
}
\caption{ RMSE (categorical only) (Pattern: General Missing, Mechanism: MNAR)}
\end{table}
\vspace{-0.3cm}

\newpage
\subsection{Downstream Task Performance: regression ($R^2$)}
We present the individual results for Downstream Task Performance, which evaluates the model's performance when trained on the imputed dataset with a newly introduced target variable (y label).

\begin{table}[!h]
\centering
\resizebox{0.3\textwidth}{!}{
\begin{tabular}{@{}c|cc|cc@{}}
\toprule
\textbf{Method} & \textbf{Rank} & \textbf{Avg} & \textbf{Diabetes} & \textbf{News} \\ \midrule
Naive & 7.6 & 30.0 & 27.1 & 32.9 \\
KNN & 9.0 & 29.4 & 26.5 & 32.2 \\
EM & \underline{5.2} & \underline{31.1} & 28.3 & 33.8 \\
MissForest & 6.0 & 30.8 & 27.6 & \underline{33.9} \\
MiceForest & 9.1 & 29.1 & 26.7 & 31.5 \\
GAIN & 7.6 & 29.8 & 26.3 & 33.4 \\
MIWAE & 7.1 & 30.0 & 27.6 & 32.4 \\
MIRACLE & 9.6 & 28.3 & 24.5 & 32.0 \\
HyperImpute & 7.2 & 29.9 & 27.5 & 32.2 \\
TDM & 5.2 & 31.0 & \underline{28.6} & 33.3 \\
IGRM & 5.7 & 27.8 & 27.8 & - \\
ReMasker & 8.0 & 29.3 & 25.4 & 33.2 \\ \midrule
\textbf{PMAE-trf} & \textbf{5.0} & \textbf{31.5} & 28.2 & \textbf{34.7} \\
\textbf{PMAE-mix} & 5.8 & 31.0 & \textbf{29.2} & 32.8 \\ \bottomrule
\end{tabular}
}
\caption{ Downstream Task Performance (regression) (Pattern: Monotone Missing, Mechanism: MNAR)}
\end{table}
\vspace{-0.3cm}

\begin{table}[!h]
\centering
\resizebox{0.3\textwidth}{!}{
\begin{tabular}{@{}c|cc|cc@{}}
\toprule
\textbf{Method} & \textbf{Rank} & \textbf{Avg} & \textbf{Diabetes} & \textbf{News} \\ \midrule
Naive & 8.2 & 23.2 & 26.3 & 20.2 \\
KNN & 7.2 & 24.0 & 27.0 & 21.0 \\
EM & 5.6 & 24.1 & 28.4 & 19.8 \\
MissForest & 5.9 & 23.9 & 26.3 & \textbf{21.5} \\
MiceForest & 9.8 & 20.5 & 25.3 & 15.8 \\
GAIN & 9.0 & 22.5 & 26.3 & 18.7 \\
MIWAE & 9.2 & 21.9 & 26.1 & 17.8 \\
MIRACLE & 10.1 & 20.6 & 23.1 & 18.0 \\
HyperImpute & 6.9 & 23.2 & 27.7 & 18.6 \\
TDM & 6.8 & 23.5 & 27.2 & 19.9 \\
IGRM & 6.1 & \textbf{27.9} & 27.9 & - \\
ReMasker & \underline{5.2} & 25.0 & \underline{28.5} & \underline{21.4} \\ \midrule
\textbf{PMAE-trf} & \textbf{4.6} & \underline{24.6} & \textbf{29.2} & 20.0 \\
\textbf{PMAE-mix} & 6.5 & 23.6 & 28.0 & 19.2 \\ \bottomrule
\end{tabular}
}
\caption{ Downstream Task Performance (regression) (Pattern: Quasi-Monotone Missing, Mechanism: MNAR)}
\end{table}
\vspace{-0.3cm}

\begin{table}[!h]
\centering
\resizebox{0.3\textwidth}{!}{
\begin{tabular}{@{}c|cc|cc@{}}
\toprule
\textbf{Method} & \textbf{Rank} & \textbf{Avg} & \textbf{Diabetes} & \textbf{News} \\ \midrule
Naive & 9.2 & 17.0 & 15.7 & 18.3 \\
KNN & 7.0 & 18.8 & 17.3 & \textbf{20.2} \\
EM & 6.2 & 19.1 & 19.3 & 18.9 \\
MissForest & 6.7 & 18.8 & 18.3 & 19.3 \\
MiceForest & 10.0 & 17.7 & 17.1 & 18.3 \\
GAIN & 8.6 & 17.9 & 16.5 & 19.2 \\
MIWAE & 7.8 & 18.7 & 18.6 & 18.7 \\
MIRACLE & 10.2 & 16.5 & 14.6 & 18.3 \\
HyperImpute & 7.2 & 18.6 & 18.0 & 19.2 \\
TDM & 7.0 & 18.9 & 19.3 & 18.4 \\
IGRM & 5.4 & 18.9 & 18.9 & - \\
ReMasker & 7.0 & 18.8 & 17.9 & \underline{19.7} \\ \midrule
\textbf{PMAE-trf} & \underline{4.4} & \underline{19.6} & \underline{20.1} & 19.1 \\ 
\textbf{PMAE-mix} & \textbf{4.0} & \textbf{19.9} & \textbf{20.4} & 19.3 \\ \bottomrule
\end{tabular}
}
\caption{ Downstream Task Performance (regression) (Pattern: General Missing, Mechanism: MNAR)}
\end{table}
\vspace{-0.3cm}

\newpage
\subsection{Downstream Task Performance: classification (AUROC)}

\begin{table}[!h]
\centering
\resizebox{0.6\textwidth}{!}{
\begin{tabular}{@{}c|cc|ccccccc@{}}
\toprule
\textbf{Method} & \textbf{Rank} & \textbf{Avg} & \textbf{Wine} & \textbf{Obesity} & \textbf{Bike} & \textbf{Shoppers} & \textbf{Letter} & \textbf{Default} & \textbf{Adult} \\ \midrule
Naive & 8.7 & 90.5 & 75.4 & 92.2 & 96.0 & 97.4 & 81.0 & 94.3 & 96.9 \\
KNN & 6.5 & 91.1 & 75.8 & 93.7 & 96.7 & 97.2 & \textbf{83.3} & 93.8 & 97.2 \\
EM & 6.8 & 91.5 & 75.7 & 93.1 & 96.4 & 98.2 & 82.2 & 96.8 & 98.4 \\
MissForest & 7.0 & 91.1 & 76.0 & 93.5 & 97.5 & 97.4 & 82.3 & 94.3 & 97.1 \\
MiceForest & 7.0 & 90.8 & 75.7 & 93.6 & 96.7 & 97.1 & 82.4 & 93.2 & 97.2 \\
GAIN & 7.5 & 91.1 & 75.5 & 91.7 & 97.9 & 97.9 & 81.5 & 95.7 & 97.8 \\
MIWAE & 9.8 & 88.9 & 75.2 & 91.1 & 95.5 & 92.1 & 81.7 & 89.9 & 97.2 \\
MIRACLE & 8.4 & 90.2 & 75.5 & 90.9 & 95.9 & 96.4 & 82.4 & 93.4 & 97.0 \\
HyperImpute & 6.6 & 91.2 & 75.8 & 92.6 & 97.5 & 97.2 & 83.2 & 94.2 & 97.6 \\
TDM & 7.0 & 91.0 & 75.9 & 93.4 & 96.2 & 97.6 & 82.1 & 94.7 & 97.4 \\
IGRM & 4.6 & 85.1 & 76.0 & \textbf{94.3} & - & - & - & - & - \\
ReMasker & 8.6 & 90.5 & 74.2 & 89.2 & \textbf{98.7} & 97.7 & 79.3 & 96.0 & 98.7 \\ \midrule
\textbf{PMAE-trf} & \textbf{4.1} & \textbf{92.4} & \underline{76.2} & \underline{93.8} & \underline{98.6} & \textbf{98.5} & \underline{83.2} & \textbf{97.3} & \underline{99.0} \\
\textbf{PMAE-mix} & \underline{4.2} & \textbf{92.4} & \textbf{76.3} & \underline{93.8} & 98.4 & \textbf{98.5} & \underline{83.2} & \textbf{97.3} & \textbf{99.1} \\ \bottomrule
\end{tabular}
}
\caption{ Downstream Task Performance (classification) (Pattern: Monotone Missing, Mechanism: MNAR)}
\end{table}
\vspace{-0.3cm}

\begin{table}[!h]
\centering
\resizebox{0.6\textwidth}{!}{
\begin{tabular}{@{}c|cc|ccccccc@{}}
\toprule
\textbf{Method} & \textbf{Rank} & \textbf{Avg} & \textbf{Wine} & \textbf{Obesity} & \textbf{Bike} & \textbf{Shoppers} & \textbf{Letter} & \textbf{Default} & \textbf{Adult} \\ \midrule
Naive & 8.6 & 89.1 & 74.4 & 92.9 & 95.8 & 92.7 & 78.2 & 93.2 & 96.8 \\
KNN & 7.2 & 89.1 & 75.3 & \underline{94.0} & 94.5 & 91.4 & \textbf{81.9} & 90.9 & 95.7 \\
EM & 5.9 & 89.6 & 75.2 & 93.3 & 93.5 & \textbf{94.1} & 79.5 & 94.2 & 97.6 \\
MissForest & 7.8 & 89.0 & 75.0 & 93.6 & 95.4 & 91.7 & 79.8 & 91.6 & 95.7 \\
MiceForest & 10.6 & 88.1 & 75.0 & 93.0 & 93.9 & 90.1 & 79.5 & 90.3 & 95.0 \\
GAIN & 7.9 & 88.9 & 74.0 & 90.9 & \underline{95.9} & 93.1 & 78.0 & 93.3 & 97.2 \\
MIWAE & 11.1 & 87.2 & 74.6 & 91.8 & 92.7 & 89.2 & 78.6 & 88.5 & 94.9 \\
MIRACLE & 7.3 & 89.7 & 74.9 & 93.0 & 95.4 & 92.7 & 80.1 & 94.4 & 97.1 \\
HyperImpute & 6.5 & 89.0 & 75.4 & 91.6 & 95.8 & 92.3 & 80.8 & 91.5 & 95.7 \\
TDM & 8.9 & 88.6 & 75.2 & 92.9 & 92.9 & 92.5 & 78.7 & 91.8 & 96.1 \\
IGRM & 3.8 & 85.0 & \textbf{75.5} & \textbf{94.5} & - & - & - & - & - \\
ReMasker & 4.3 & 90.2 & 75.4 & 93.5 & 95.8 & 93.9 & 80.3 & 94.7 & 97.9 \\ \midrule
\textbf{PMAE-trf} & \textbf{3.1} & \textbf{90.5} & \textbf{75.5} & 93.9 & \underline{95.9} & 94.0 & \underline{81.0} & \textbf{94.8} & \textbf{98.3} \\
\textbf{PMAE-mix} & \underline{3.8} & \underline{90.4} & 75.3 & 93.7 & \textbf{96.1} & \textbf{94.1} & 80.9 & \textbf{94.8} & \underline{98.2} \\ \bottomrule
\end{tabular}
}
\caption{ Downstream Task Performance (classification) (Pattern: Quasi-Monotone Missing, Mechanism: MNAR)}
\end{table}
\vspace{-0.3cm}

\begin{table}[!h]
\centering
\resizebox{0.6\textwidth}{!}{
\begin{tabular}{@{}c|cc|ccccccc@{}}
\toprule
\textbf{Method} & \textbf{Rank} & \textbf{Avg} & \textbf{Wine} & \textbf{Obesity} & \textbf{Bike} & \textbf{Shoppers} & \textbf{Letter} & \textbf{Default} & \textbf{Adult} \\ \midrule
Naive & 7.8 & 85.3 & 72.6 & 88.0 & 91.8 & 90.0 & 74.6 & 90.0 & 90.4 \\
KNN & 6.9 & 85.3 & 73.2 & 89.5 & 90.8 & 88.7 & \textbf{78.0} & 86.5 & 90.0 \\
EM & 6.0 & 86.1 & 73.2 & 88.5 & 90.0 & 90.8 & 76.1 & 91.8 & 92.0 \\
MissForest & 7.3 & 85.3 & 72.8 & 88.8 & 92.6 & 89.0 & 76.5 & 87.8 & 89.3 \\
MiceForest & 11.0 & 83.2 & 71.5 & 86.1 & 90.5 & 85.9 & 75.0 & 85.0 & 88.1 \\
GAIN & 8.3 & 84.3 & 70.4 & 83.9 & 91.7 & 90.4 & 71.5 & 91.2 & 90.8 \\
MIWAE & 11.5 & 82.5 & 72.6 & 86.2 & 89.0 & 83.8 & 74.4 & 84.7 & 87.2 \\
MIRACLE & 7.2 & 85.5 & 72.8 & 87.4 & 91.9 & 89.0 & 76.1 & 89.4 & 91.8 \\
HyperImpute & 6.7 & 84.1 & 73.1 & \underline{91.0} & 91.9 & 77.5 & 76.6 & 88.9 & 90.0 \\
TDM & 9.2 & 84.5 & 73.0 & 87.5 & 89.9 & 88.9 & 75.4 & 88.3 & 88.8 \\
IGRM & 2.7 & 83.3 & \underline{73.8} & \textbf{92.8} & - & - & - & - & - \\
ReMasker & 4.9 & 86.6 & 72.9 & 88.9 & 91.3 & 91.0 & 76.6 & 92.0 & 93.1 \\ \midrule
\textbf{PMAE-trf} & \textbf{2.8} & \textbf{87.6} & 73.7 & 90.1 & \textbf{93.0} & \textbf{91.1} & \underline{77.8} & \textbf{92.5} & \textbf{94.8} \\
\textbf{PMAE-mix} & \underline{3.2} & \underline{87.4} & \textbf{73.9} & 89.7 & \underline{92.7} & \textbf{91.1} & \underline{77.8} & \underline{92.4} & \underline{94.6} \\ \bottomrule
\end{tabular}
}
\caption{ Downstream Task Performance (classification) (Pattern: General Missing, Mechanism: MNAR)}
\end{table}
\vspace{-0.3cm}

\newpage
\section{Additional Result: MCAR / MAR / MNAR in the Monotone Missing Pattern}
Although MNAR is applied to all settings, we further verify the correct application of the missing mechanisms by analyzing performance differences between various algorithms. The significant performance drops observed across all methods confirm that MNAR is the most challenging setting. Consequently, we focus our evaluation on this most difficult scenario.

\begin{figure*}[!h]
    \centering
        \subfigure[Imputation Accuracy (Avg)]{\includegraphics[width=0.62\textwidth]{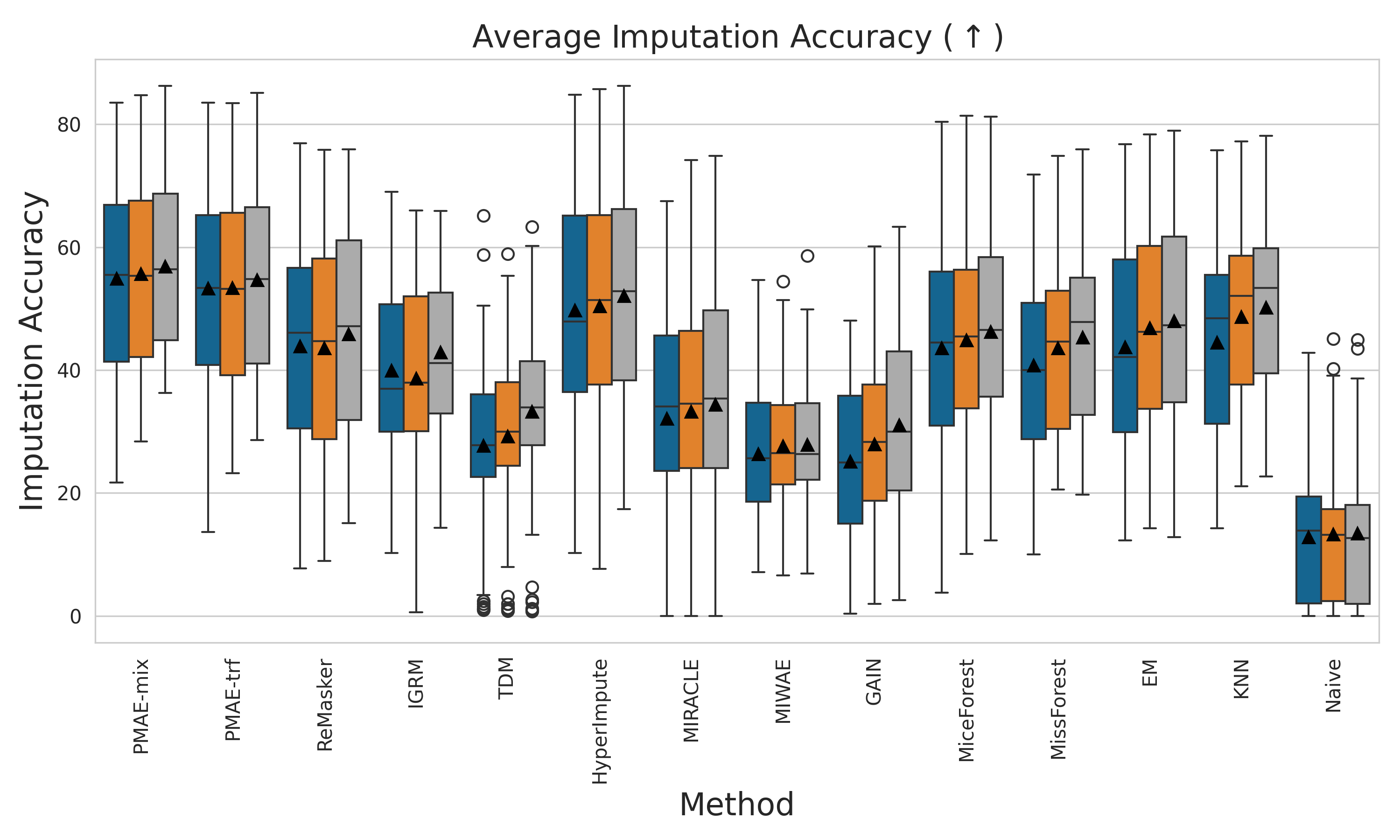}} \hfill%
        \subfigure[Imputation Accuracy (Rank)]{\includegraphics[width=0.37\textwidth]{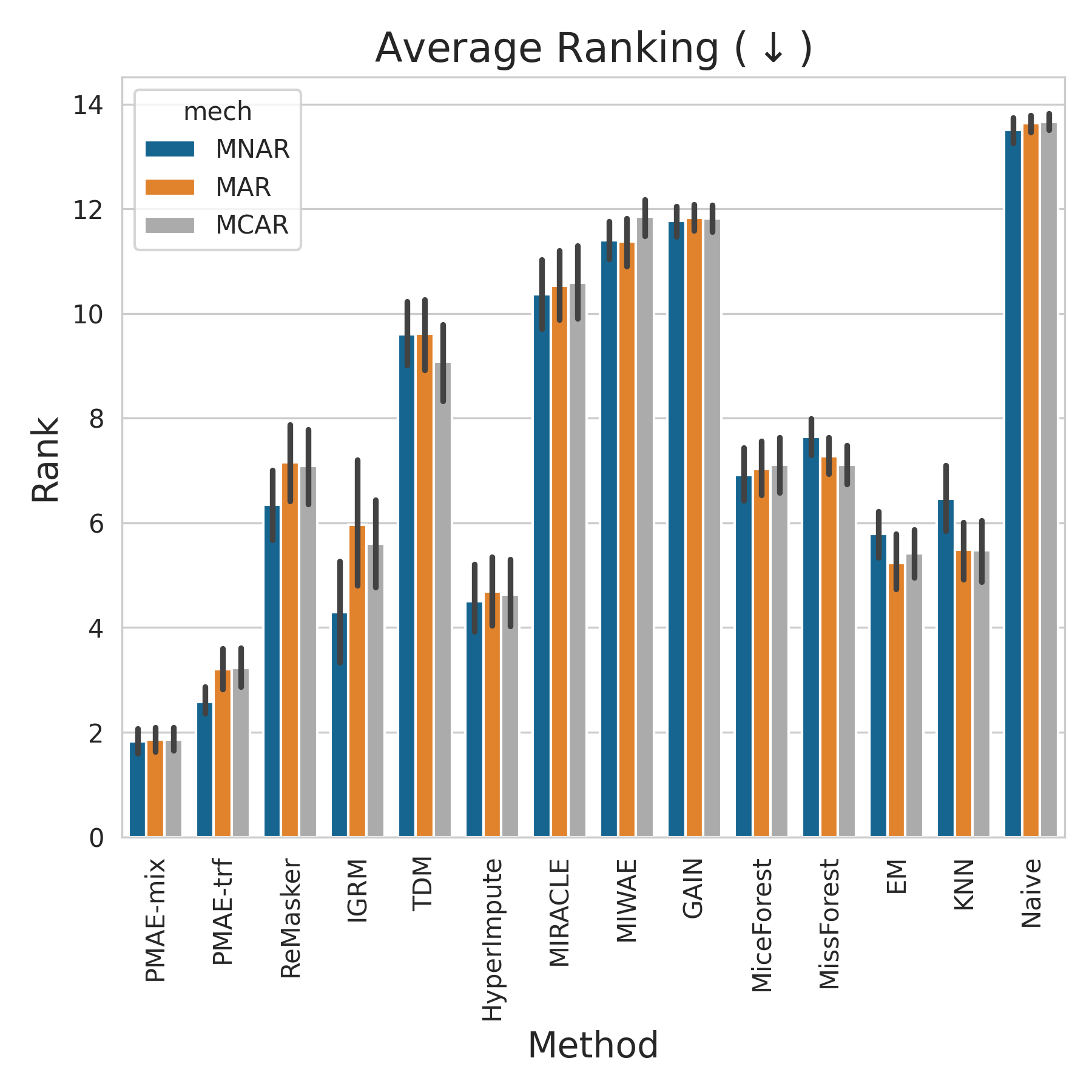}} \hfill%
\end{figure*} 

\end{document}